\documentclass[10pt,twocolumn,letterpaper]{article}

\usepackage[pagenumbers]{iccv} % To force page numbers, e.g. for an arXiv version

\definecolor{iccvblue}{rgb}{0.21,0.49,0.74}
\usepackage[pagebackref,breaklinks,colorlinks,allcolors=iccvblue]{hyperref}

\usepackage{algorithm}
\usepackage{listings}
\usepackage{graphicx}
\usepackage{amsmath}
\usepackage{amssymb}
\usepackage{booktabs}
\usepackage{tabularx}
\usepackage{makecell}
\usepackage{enumitem}
\usepackage{multirow}

\def\paperID{10500} % *** Enter the Paper ID here
\def\confName{ICCV}
\def\confYear{2025}

\title{Diffusion Model with Perceptual Loss}

\author{Shanchuan Lin \quad Xiao Yang \\
ByteDance Inc.\\
{\tt\small \{peterlin,yangxiao.0\}@bytedance.com}
}

\newcommand\banner{%
    \vspace{3pt}
    \centering
    \footnotesize
    \setlength\tabcolsep{1.5pt}
    \begin{tabular}{ c c }
        \includegraphics[width=0.495\textwidth]{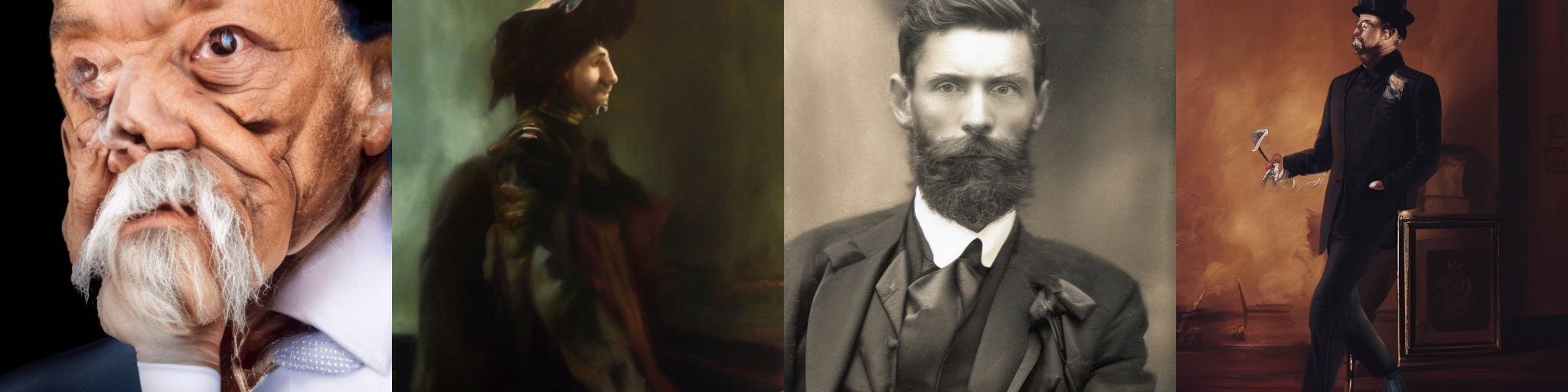} & 
        \includegraphics[width=0.495\textwidth]{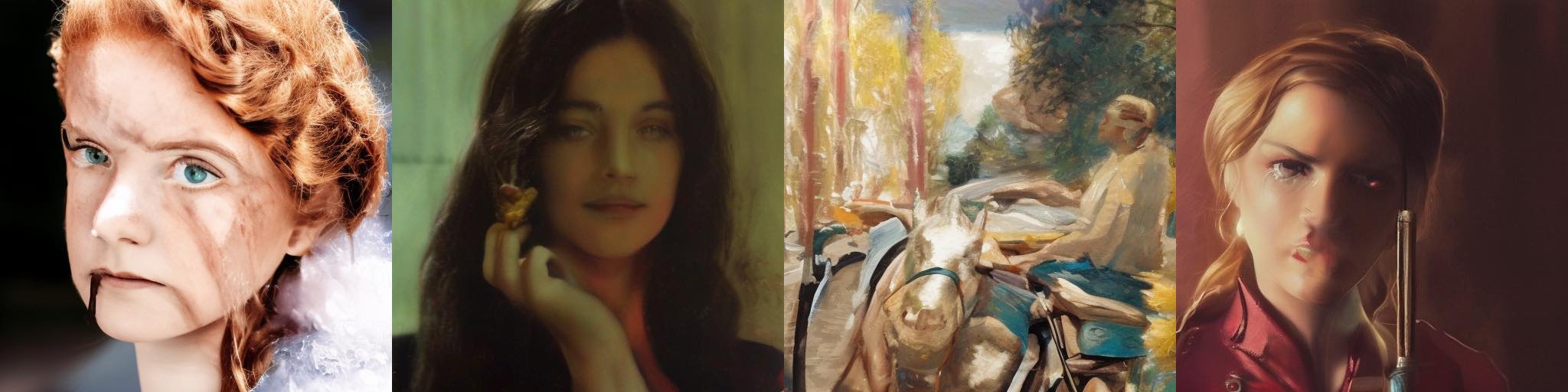} \\
        \includegraphics[width=0.495\textwidth]{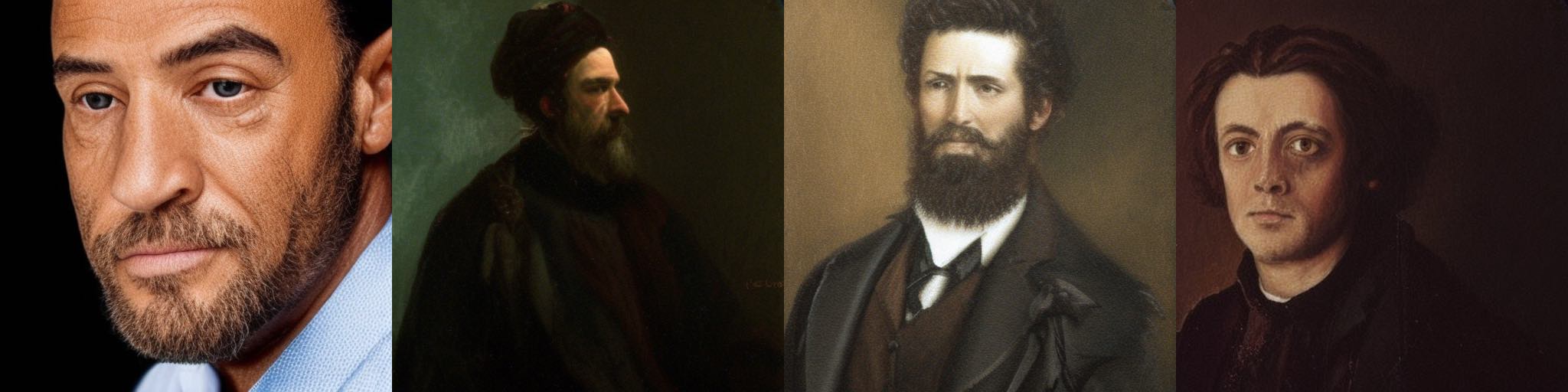} & 
        \includegraphics[width=0.495\textwidth]{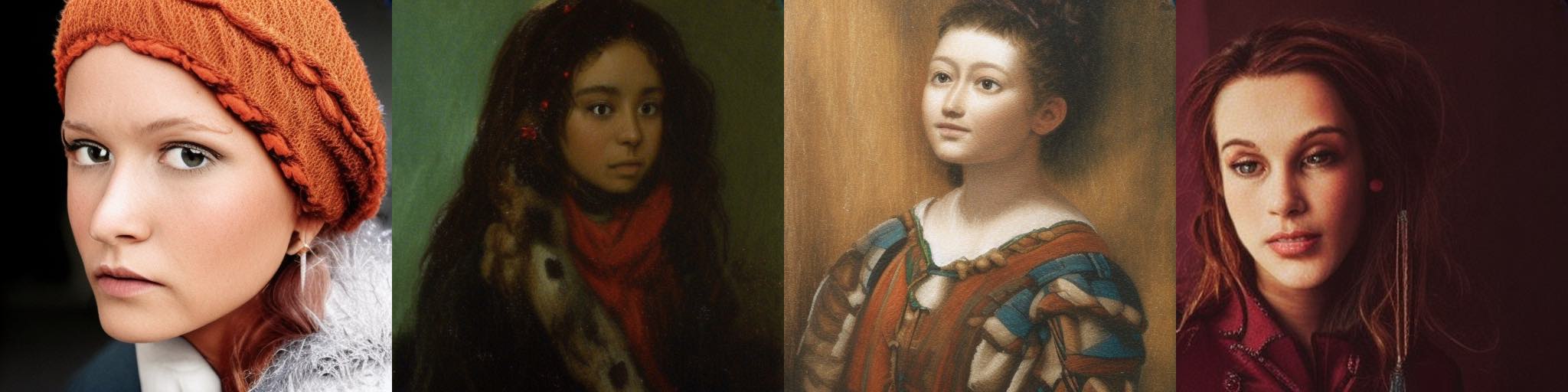} \\
        A handsome man & A beautiful girl \\
    \end{tabular}
    \vspace{-6pt}
    \captionof{figure}{The diffusion model trained with MSE loss generates unrealistic samples without guidance (top row). Our proposed self-perceptual loss can generate realistic samples without guidance. The loss objective is important in shaping the learned distribution.}
    \label{fig:banner}
    \vspace{-12pt}
}

\begin{document}
\maketitle
\begin{abstract}
Diffusion models without guidance generate very unrealistic samples. Guidance is used ubiquitously, and previous research has attributed its effect to low-temperature sampling that improves quality by trading off diversity. However, this perspective is incomplete. Our research shows that the choice of the loss objective is the underlying reason raw diffusion models fail to generate desirable samples. In this paper, (1) our analysis shows that the loss objective plays an important role in shaping the learned distribution and the MSE loss derived from theories holds assumptions that misalign with data in practice; (2) we explain the effectiveness of guidance methods from a new perspective of perceptual supervision; (3) we validate our hypothesis by training a diffusion model with a novel self-perceptual loss objective and obtaining much more realistic samples without the need for guidance. We hope our work paves the way for future explorations of the diffusion loss objective.
\end{abstract}

\let\thefootnote\relax\footnotetext{Model: \href{https://huggingface.co/ByteDance/sd2.1-base-zsnr-laionaes6-perceptual}{hf.co/ByteDance/sd2.1-base-zsnr-laionaes6-perceptual}}

\vspace{-1em}
\section{Introduction}

Conceptually, diffusion models \cite{ho2020denoising,sohldickstein2015deep,song2021scorebased} work by transforming noise to data samples through repeated denoising. Formally, each denoising step can be viewed from the lens of score matching \cite{song2021scorebased} such that the model learns to predict the drift (score) of a stochastic differential equation (SDE), or equivalently the gradient (flow) of an ordinary differential equation (ODE), that transports samples from one distribution (noise) to another distribution (image, video, \etc) \cite{liu2022flow}.

Diffusion models are commonly parameterized as neural networks and the training objective minimizes the squared distance between the model prediction and the target score through stochastic gradient descent \cite{JMLR:v6:hyvarinen05a}. This is also commonly referred to as the mean squared error (MSE) loss. 

Although diffusion models are supposed to transport samples from the noise to the data distribution by theory, samples generated by diffusion models without guidance are often of poor quality as shown in \cref{fig:banner}, despite the improvements in model architecture \cite{chen2023pixartalpha,karras2023analyzing,peebles2023scalable,podell2023sdxl,saharia2022photorealistic,ramesh2022hierarchical,rombach2022highresolution,esser2024scalingrectifiedflowtransformers,liu2024playgroundv3improvingtexttoimage}, formulation \cite{lipman2023flow,liu2022flow,karras2022elucidating}, and sampling strategy \cite{song2022denoising,lu2022dpmsolver,lu2023dpmsolver,karras2022elucidating}.

Classifier-free guidance (CFG) \cite{ho2022classifierfree} is applied almost ubiquitously in state-of-the-art diffusion models across modalities to improve sample quality, \eg, text-to-image \cite{podell2023sdxl,rombach2022highresolution,saharia2022photorealistic,chen2023pixartalpha,ramesh2022hierarchical,zheng2022movq,pernias2023wuerstchen}, text-to-video \cite{guo2023animatediff,blattmann2023align,zhou2023magicvideo,ho2022imagen,singer2022makeavideo,blattmann2023stable}, text-to-3d \cite{shi2023mvdream,wang2023prolificdreamer,poole2022dreamfusion}, image-to-video \cite{blattmann2023stable,yan2023magicprop}, video-to-video \cite{chang2023magicdance,esser2023structure}, \etc Previous research has attributed its effect to low-temperature sampling \cite{ho2022classifierfree}, as if the quality improvement is a result of trading off diversity. However, our research provides a different perspective.

In this paper, we seek to uncover the fundamental cause of why diffusion models without guidance fail to generate desirable samples. Our analysis suggests that the loss objective plays an important role in shaping the learned distribution, and the common MSE loss objective derived from theories holds assumptions that misalign with data in practice. Based on these findings, we experiment using a perceptual distance loss objective. Specifically, we propose a novel self-perceptual objective that uses the diffusion model itself as the perceptual loss. Our method generates much more realistic samples without guidance. Our main contributions are as follows:

\begin{itemize}
  \item Our analysis uncovers the important effect of the loss objective in shaping the learned probability distribution of diffusion models, and shows that the MSE loss holds assumptions that misalign with data in practice. More importantly, we show that the loss objective is open for exploration without a single theoretically correct solution.

  \item We provide a different perspective on guidance methods through the lens of perceptual supervision instead of low-temperature sampling.
  
  \item To the best of our knowledge, we are the first to apply perceptual loss to diffusion training. We propose a novel self-perceptual loss that uses the diffusion model itself as the perceptual network. We demonstrate its effectiveness in improving sample quality.
\end{itemize}

Our work studies the underlying cause of why diffusion models generate poor samples without guidance. We hope our work paves the way for more future explorations in the diffusion loss objective.
\section{Related Work}

Guidance methods alter the model prediction and guide the sample toward desired regions during the generation process. \textbf{Classifier Guidance} \cite{dhariwal2021diffusion} adds classifier gradients to the predicted score to guide the sample generation to maximize the classification. It can turn an unconditional diffusion model conditional. However, it is not evident why applying classifier guidance on an already conditional diffusion model can significantly improve sample quality. Previous research has attributed it to low-temperature sampling \cite{ho2022classifierfree,karras2024guidingdiffusionmodelbad}. \textbf{Classifier-Free Guidance (CFG)} \cite{ho2022classifierfree} uses Bayes’ rule and finds that the diffusion model itself can be inverted as an implicit classifier. Specifically, the model is queried both conditionally and unconditionally at every inference step and the difference is amplified toward the conditional direction. Both methods only work for conditional generation and entangle sample quality with conditional alignment \cite{karras2024guidingdiffusionmodelbad}. \textbf{Self-Supervised Guidance} \cite{Hu2023GuidedDF} uses self-supervised networks to generate synthetic clustering labels for unconditional data. This allows unconditional data to use CFG for improving quality. \textbf{Guidance-Free Training} \cite{chen2025visualgenerationguidance} shows that CFG can be applied at training. This bakes the guided flow into the model and saves computation during inference. More recently, \textbf{Discriminator Guidance} \cite{Kim2022RefiningGP} proposes to train a discriminator network to classify real and generated samples and use it as guidance during diffusion generation. \textbf{Self-Attention Guidance} \cite{Hong2022ImprovingSQ} finds that the self-attention map of the diffusion model can be exploited to enhance quality. \textbf{Autoguidance} \cite{karras2024guidingdiffusionmodelbad} finds a smaller or less-trained model can be used as negative guidance to improve quality. Although these methods are effective in improving quality, they also present issues such as increased complexity and reduced diversity \cite{ho2022classifierfree,karras2024guidingdiffusionmodelbad}, \etc. On the other hand, our research aims to explore the underlying issue: why diffusion models without guidance fail in the first place.

The loss objective of diffusion models has been studied by prior works. Loss weighting is found to influence perceptual quality and likelihood evaluation \cite{Choi2022PerceptionPT,Song2021MaximumLT,Hang2023EfficientDT} but still cannot produce good samples without guidance. Multi-scale loss \cite{Hoogeboom2023simpleDE} is proposed to improve high-resolution generation. Smoothness penalty \cite{Guo2023SmoothDC} is proposed to enforce smoother latent traversal. $l1$ distance is explored for colorization and in-painting tasks \cite{Saharia2021PaletteID}. The squared distance objective is almost ubiquitously adopted. Perceptual loss has been explored in consistency models \cite{song2023consistencymodels}, but we are the first to explore perceptual loss in diffusion models.

In recent years, the size of diffusion models has increased to multi-billion parameters \cite{esser2024scalingrectifiedflowtransformers,liu2024playgroundv3improvingtexttoimage}. The architecture has improved from convolution \cite{rombach2022highresolution,podell2023sdxl,saharia2022photorealistic} to transformers \cite{esser2024scalingrectifiedflowtransformers,liu2024playgroundv3improvingtexttoimage,chen2023pixartalpha}. More accurate solvers \cite{lu2022dpmsolver,lu2023dpmsolver} and better formulations \cite{liu2022flow,lipman2023flow,karras2022elucidating} have been proposed. However, diffusion models still generate poor samples without guidance.
\section{Analysis}

In this section, we analyze why diffusion models without guidance generate poor samples. We show that the loss objective is important in shaping the learned distribution and the MSE loss is not optimal.

\subsection{The Effect of the Loss Objective}
\label{sec:loss-objective-effect}

\begin{figure}[h]
    \centering
    \small
    \begin{subfigure}{0.24\linewidth}
        \includegraphics[width=\textwidth]{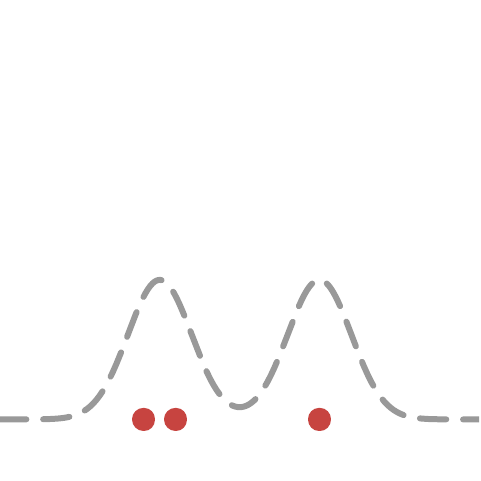}
        \caption{Actual}
        \label{fig:distribution-original}
    \end{subfigure}
    \hfill
    \begin{subfigure}{0.24\linewidth}
        \includegraphics[width=\textwidth]{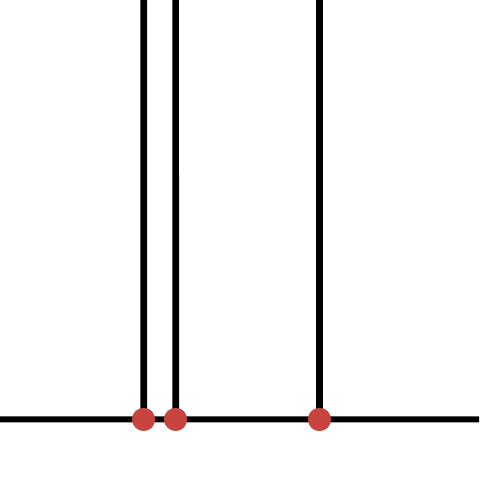}
        \caption{MLE}
        \label{fig:distribution-mle}
    \end{subfigure}
    \hfill
    \begin{subfigure}{0.24\linewidth}
        \includegraphics[width=\textwidth]{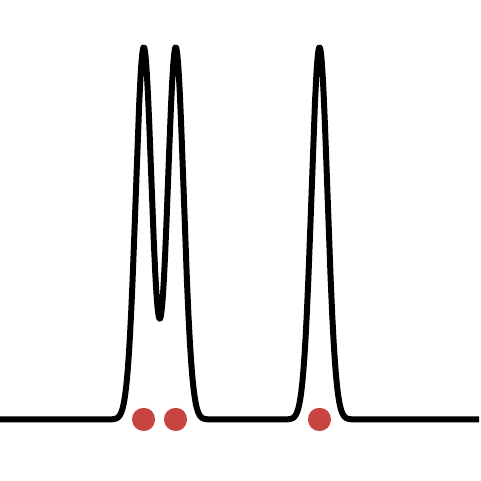}
        \caption{Learned 1}
        \label{fig:distribution-learned1}
    \end{subfigure}
    \hfill
    \begin{subfigure}{0.24\linewidth}
        \includegraphics[width=\textwidth]{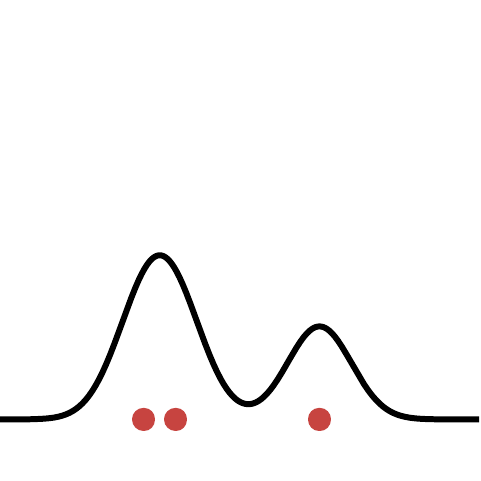}
        \caption{Learned 2}
        \label{fig:distribution-learned2}
    \end{subfigure}
    \caption{A motivating example where data samples are given and the actual distribution is unknown. Diffusion models learn the maximum likelihood estimation (MLE) distribution as the target. Neural networks create smoothness and generalization. The loss objective influences the shape of the learned distribution and can be designed with inductive biases to better drive it toward the actual distribution.}
    \label{fig:distribution-example}
    \vspace{-1em}
\end{figure}

In \cref{fig:distribution-original}, consider a simple toy scenario where some finite observed data samples (red dots) are given, and a generative model is tasked to model the unknown underlying distribution (dashed line). The problem is inherently ill-posed because any distribution with nonzero probabilities over the observed samples is a valid solution.

A special solution, the maximum likelihood estimation (MLE) distribution, only assigns probability over the observed samples and zero probability everywhere else. Its density function can be expressed as a sum of Dirac delta functions as illustrated in \cref{fig:distribution-mle}. Notice that the MLE distribution is always spiky regardless of the dataset size as long as the samples are finite because having more samples will just result in closer spikes. The MLE distribution overfits and only reproduces the observed samples at test time.

For diffusion training, the target probability flow is unique and pre-determined by the forward diffusion process \cite{song2021scorebased}. The target flow transports between the noise distribution and the MLE data distribution \cite{song2021scorebased,Song2021MaximumLT}. This means that with sufficient model capacity, the model will learn the MLE data distribution and overfit to only generate the observed samples. In practice, models are parameterized by neural networks with finite capacity, which smooths the prediction and provides desired generalization, resulting in distributions like \cref{fig:distribution-learned1,fig:distribution-learned2}. 

The shape of the learned distribution are influenced by the model capacity and the loss objective. In particular, \textbf{the loss objective can be better designed with inductive biases to drive the shape of the learned distribution toward the actual distribution}. This is often overlooked by existing research.

\subsection{The Problem of the Squared Distance}
\label{sec:analysis-squared-distance}

The squared distance was originally proposed from theoretical derivations. From the diffusion theory perspective, the reverse generative process has the same functional form as the forward Gaussian diffusion process \cite{sohldickstein2015deep}. KL divergence was chosen to measure the Gaussian divergence between the model prediction and the posterior ground truth \cite{ho2020denoising}. The negative log-likelihood for Gaussians can be further simplified to the squared distance. From the score matching perspective \cite{JMLR:v6:hyvarinen05a} and the most recent flow matching perspective \cite{lipman2023flow}, the squared distance seems to be chosen out of convenience.

First, KL divergence is not the only valid choice of divergence measurement, so the MSE loss is not the only valid choice from the theoretical viewpoint. Second, simply extending it to high-dimensional data (images) assumes the independence of each dimension (pixels), which is generally not true.

\begin{figure}[t]
    \centering
    \begin{subfigure}[b]{0.49\linewidth}
        \centering
        \includegraphics[width=\textwidth]{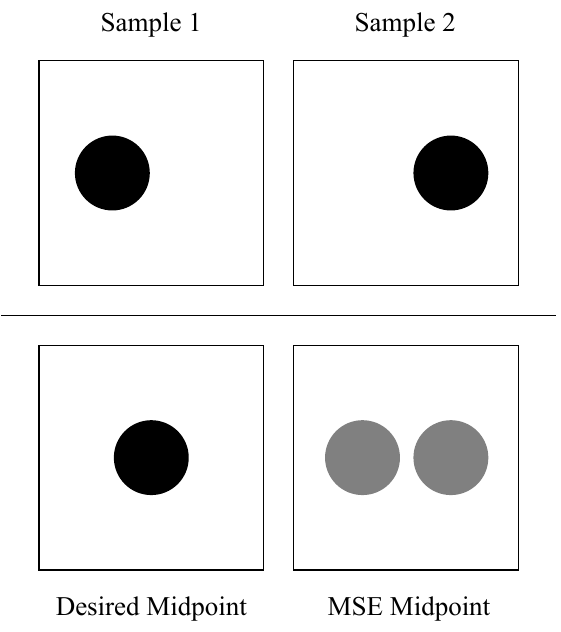}
        \caption{}
        \label{problem:distribution_dots}
    \end{subfigure}
    \begin{subfigure}[b]{0.49\linewidth}
        \centering
        \includegraphics[width=\textwidth]{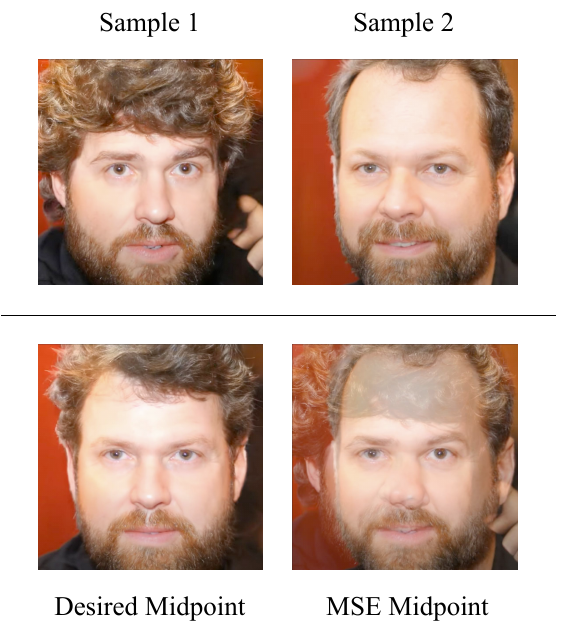}
        \caption{}
        \label{problem:distribution_faces}
    \end{subfigure}
    \vspace{-1em}
    \caption{The midpoint sample is derived by minimizing the distance to known samples by the given distance function. MSE midpoint is out-of-distribution.}
    \label{fig:midpoint}
    \vspace{-1em}
\end{figure}

\Cref{fig:midpoint} illustrates that the per-pixel MSE objective is a poor distance function for images. For example, given finite images of human faces, the underlying distribution is ambiguous to the model. The loss objective dictates the shape of the learned distribution. We would like the model to produce a distribution of semantic morphing of new faces, but MSE leads the model to learn a distribution of pixel-wise blending. This is why diffusion models with MSE loss produce out-of-distribution samples.

Even if the diffusion models employ a convolution or attention architecture that takes all the pixels into consideration, this only helps the model to predict more accurate pixel-independent MSE midpoints because it is trained to do so. The MSE objective is problematic in the first place.

\subsection{The Effect of the Guidance}

Prior research has shown that deep neural networks trained on discriminative tasks can capture the semantics of the data. Specifically, classifier networks trained on images can measure the semantic image distance better aligned with human perception \cite{zhang2018unreasonable}.

This explains the effectiveness of guidance in improving diffusion generation quality. In the case of classifier guidance \cite{dhariwal2021diffusion}, the classifier network captures the image semantics and guides the generation toward perceptually realistic samples as a side effect of maximizing the classification. For example, when the classifier is asked to classify human faces, it will assign high scores to samples with semantically correct faces with only two eyes and will penalize the pixel-blending samples that may have four eyes. Classifier-free guidance \cite{ho2022classifierfree}, derived using Bayes' rule and using the diffusion model itself as an implicit classifier, can explained for the same reason. Additionally, Generative Adversarial Networks (GANs) \cite{goodfellow2014generative} do not exhibit the pathology because the discriminator, being a deep neural network, learns to better capture the data semantics. Discriminator guidance can be explained in the same way.

We do believe low-temperature sampling is indeed a factor, as maximizing the classifier score eliminates the low-likelihood samples. However, in our perspective, we believe the MSE loss objective is the main reason diffusion models fail to generate desirable distributions without guidance, and guidance methods provide perceptual supervision in the sampling process.
\section{Method}

To validate our analysis, we experiment with directly incorporating perceptual loss into diffusion training. In \cref{sec:solution-background} we introduce the diffusion background and our model formulation. In \cref{sec:solution-self}, we propose a novel self-perceptual objective and show that the diffusion model itself can be used as a perceptual network to provide meaningful perceptual loss.

\subsection{Background}
\label{sec:solution-background}

We follow the setup of Stable Diffusion, a latent diffusion model \cite{rombach2022highresolution}. Given image latent sample $x_0 \sim \pi_0$, noise sample $\epsilon \sim \mathcal{N}(0, \textbf{I})$, and time $t \sim \mathcal{U}(1, T)$, where $t \in \mathbb{Z}, T=1000$, the forward diffusion process is defined as:
\begin{equation}
    x_t = \mathbf{forward}(x_0, \epsilon, t) = \sqrt{\bar{\alpha}_t}x_0 + \sqrt{1 - \bar{\alpha}_t}\epsilon.
    \label{eq:solution-background-xt}
\end{equation}
We use diffusion schedule with zero terminal SNR \cite{lin2023common}. The specific $\bar\alpha_t$ values are defined in \cite{lin2023common}.

Our neural network $f_\theta: \mathbb{R}^d \rightarrow \mathbb{R}^d$ is conditioned on text prompt $c$ and uses the $v$-prediction formulation \cite{salimans2022progressive,lin2023common}:
\begin{align}
    v_t &= \sqrt{\bar\alpha_t}\epsilon - \sqrt{1 - \bar\alpha_t}x_0, \\
    \hat{v}_t &= f_\theta(x_t, t, c).
\end{align}

The original MSE objective is defined as:
\begin{equation}
    \mathcal{L}_{mse} = \|\hat{v}_t - v_t\|_2^2.
    \label{eq:solution-mse-loss}
\end{equation}

\subsection{Self-Perceptual Objective}
\label{sec:solution-self}

We propose a self-perceptual objective that utilizes the MSE diffusion model itself as the perceptual network. This is not surprising as the classifier-free guidance \cite{ho2022classifierfree} also exploits the MSE diffusion model itself to provide meaning perceptual guidance at inference. 

The intuition is that even though the model's MSE flow prediction is not ideal, the model is still trained with good semantic understanding in order to predict accurate MSE flow. Therefore, our approach employs the MSE pre-trained diffusion model as our perceptual network and computes distance on its feature maps following prior perceptual loss research \cite{zhang2018unreasonable}.

% Our approach is to employ the MSE pre-trained diffusion model as our perceptual network and compute distance on its feature maps. The model is suitable for our task because it is trained on the target dataset, on the latent space, and on all noise levels $x_t$.

Specifically, we copy and freeze the diffusion model trained with the MSE loss, and we modify the architecture to return the hidden feature at layer $l$. We denote this frozen perceptual network as $p^l_*$, and the online trainable network as $f_\theta$.

During training, we sample $x_0 \sim \pi_0$, $\epsilon \sim \mathcal{N}(0, \textbf{I})$, $t \sim \mathcal{U}(1, T)$ and obtain $x_t$ through forward diffusion:
\begin{equation}
    x_t = \mathbf{forward}(x_0, \epsilon, t).
\end{equation}

We use online network $f_\theta$ to predict $\hat{v}$ and convert the prediction to $\hat{x}_0$ and $\hat{\epsilon}$:
\begin{align}
    \hat{v}_t &= f_\theta(x_t, t, c), \\
    \hat{x}_0 &= \sqrt{\bar\alpha_t}x_t - \sqrt{1-\bar\alpha_t} \hat{v},
    \label{eq:solution-background-predx} \\
    \hat{\epsilon} &= \sqrt{\bar\alpha_t} \hat{v} + \sqrt{1-\bar\alpha_t} x_t.
    \label{eq:solution-background-predeps}
\end{align}

Then, we sample a new timestep $t' \sim \mathcal{U}(1, T)$, and compute the ground-truth $x_{t'}$ and the predicted $\hat{x}_{t'}$ through forward diffusion:
\begin{align}
    x_{t'} &= \textbf{forward}(x_0, \epsilon, t'),\\
    \hat{x}_{t'} &= \textbf{forward}(\hat{x}_0, \hat\epsilon, t').
\end{align}

Notice that the $\hat{x}_t'$ derived here is equivalent to a single DDIM solver step from timestep $t$ to $t'$.

Finally, we pass both of them through the frozen perceptual network $p^l_*$ and compute the distance on its hidden feature at layer $l$. We find only using the hidden feature at the midblock layer yields the best result. We refer to our method as the Self-Perceptual (SP) objective:
\begin{equation}
    \mathcal{L}_{sp} = \| p^l_*(\hat{x}_{t'}, t', c) - p^l_*(x_{t'}, t', c) \|_2^2,
\end{equation}
\noindent
The code is provided in the supplementary materials.

We highlight that this design intentionally avoids introducing any external components, allowing us to isolate the study on loss objective. It is also practical, as we can easily finetune existing MSE diffusion models to SP using itself.
\section{Evaluation}

We first finetune Stable Diffusion v2.1 \cite{rombach2022highresolution} using our formulation and MSE loss $\mathcal{L}_{mse}$ on a LAION aesthetic 6+ dataset \cite{schuhmann2022laion5b} for 60k iterations. This ensures that the training data is consistent for fair comparisons. We use 10\% conditional dropout to support CFG for evaluation comparison. Then we copy and freeze the MSE model as our perceptual network, and continue training the online network with our self-perceptual objective $\mathcal{L}_{sp}$ for 50k iterations. We use learning rate 3e-5, batch size 896, and EMA decay 0.9995. We verify both networks are trained till convergence for a fair comparison.

We also train unconditional models following the same procedure except we always use an empty prompt during training and inference. This is to demonstrate our approach also works for unconditional generation.

For inference, we use deterministic DDIM sampler \cite{song2022denoising}, and make sure the sampler correctly starts from the zero terminal SNR at the last timestep $T$ \cite{lin2023common}.

\subsection{Qualitative}
\label{sec:evaluation-qualitative}

\Cref{fig:qualitative} shows the conditional generation results. Our self-perceptual objective has significant quality improvement over the MSE objective. This validates our analysis in \cref{sec:loss-objective-effect,sec:analysis-squared-distance} that the MSE loss is the cause for poor generation quality and a better loss objective such as the perceptual distance can indeed improve quality.

Notice that the results of the MSE and the self-perceptual objective share very similar content and layout when generated from the same initial noise. 
As stated in \cref{sec:loss-objective-effect}, this is because the underlying target MLE probability flow is exactly the same, and changing the loss objective only influences the model's learned generalization, so the same noise will map to data in the similar region. On the other hand, CFG changes the flow drastically.

Compared to CFG, the self-perceptual objective only affects sample quality but not text alignment. This is especially evident in \cref{fig:evaluation-qualitative-dog}, where CFG enhances the text condition, while our SP model and the original MSE model are more aligned with the training dataset distribution, where LAION dataset has less-aligned dirty captions.

\Cref{fig:evaluation-qualitative-painting} shows the negative artifact of CFG. The model has already overfitted the image to the very specific prompt, and the high CFG scale causes unnatural artifacts. Our self-perceptual objective does not suffer from this issue.

\begin{figure}[h]
    \centering
    \small
    \begin{tabularx}{\linewidth}{|X|X}
        MSE & Self-Perceptual
    \end{tabularx}
    \begin{subfigure}[b]{0.495\linewidth}
        \centering
        \includegraphics[width=\textwidth]{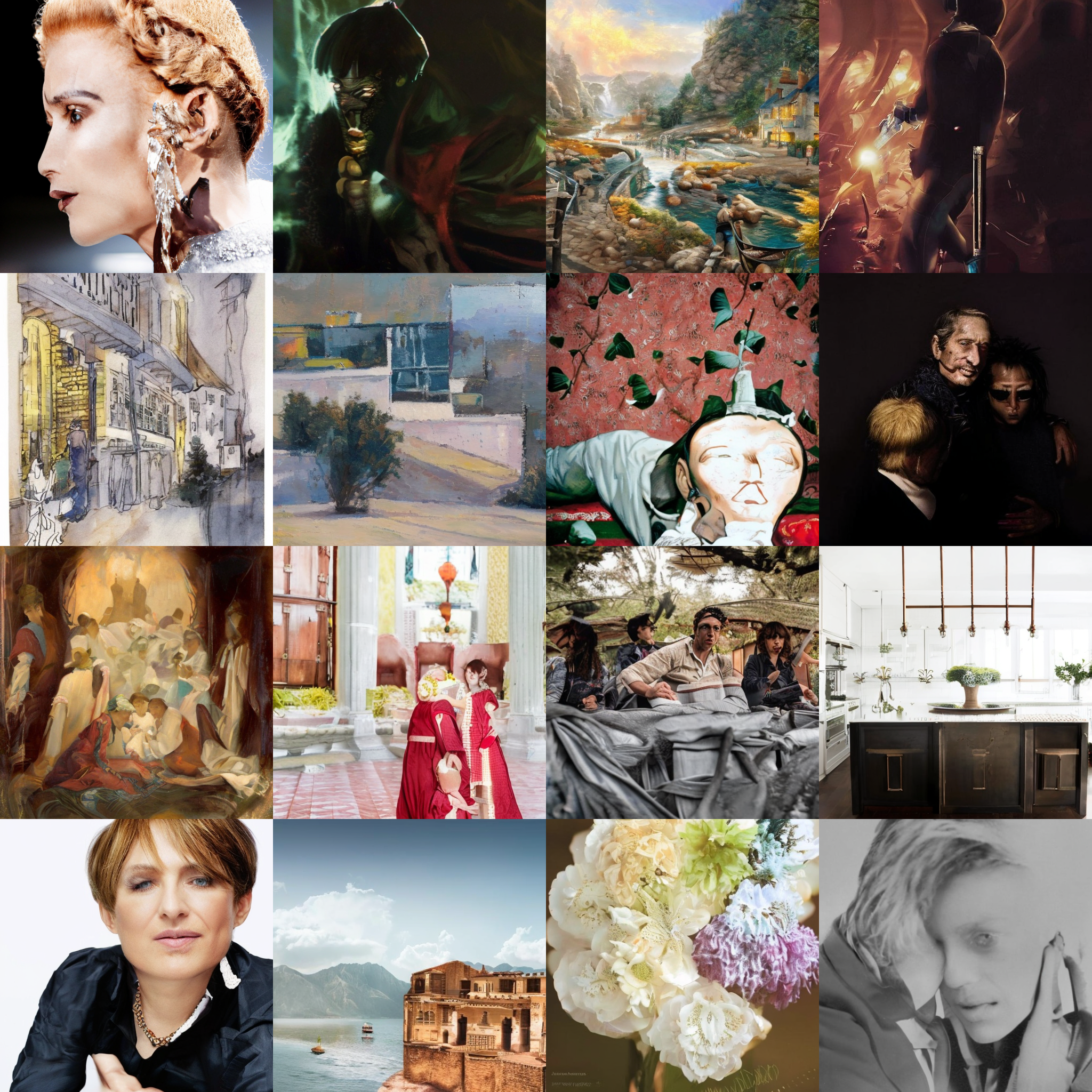}
    \end{subfigure}
    \begin{subfigure}[b]{0.495\linewidth}
        \centering
        \includegraphics[width=\textwidth]{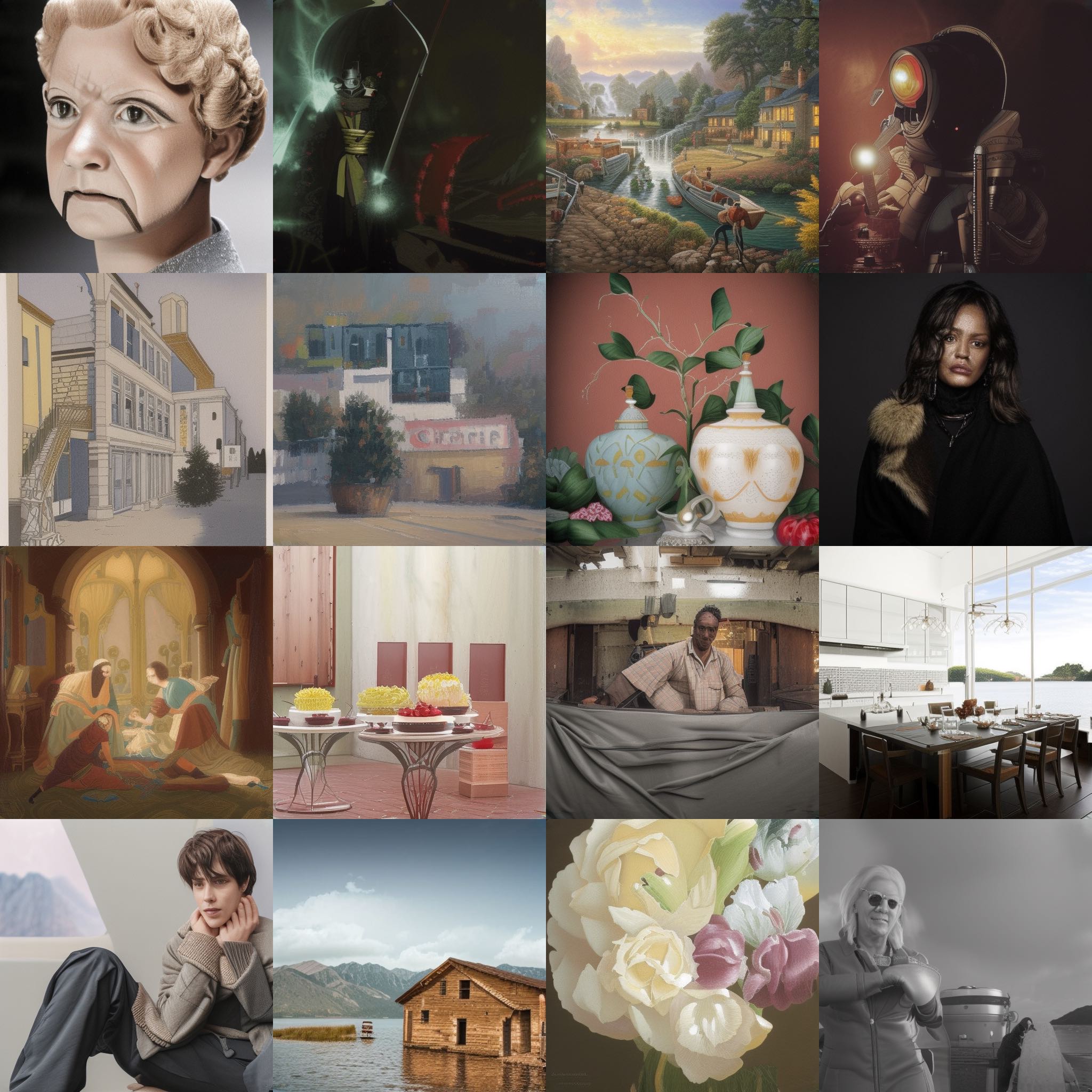}
    \end{subfigure}
    \vspace{-2em}
    \caption{Unconditional generation. Both use DDIM 1000 steps with the same seed. Our self-perceptual objective can improve unconditional generation quality.}
    \label{fig:evaluation-qualitative-uncond}
    % \vspace{-2em}
\end{figure}

\Cref{fig:evaluation-qualitative-uncond} shows that our approach can also improve sample quality for unconditional generation.

\begin{figure*}[t]
    \centering
    \captionsetup{justification=raggedright,singlelinecheck=false}
    \small
    \begin{tabularx}{\textwidth}{|X|X|X|X|X|X}
        MSE & Self-Perceptual & MSE + CFG & MSE & Self-Perceptual & MSE + CFG
    \end{tabularx}
    \begin{subfigure}[b]{0.495\textwidth}
        \centering
        \includegraphics[width=\textwidth]{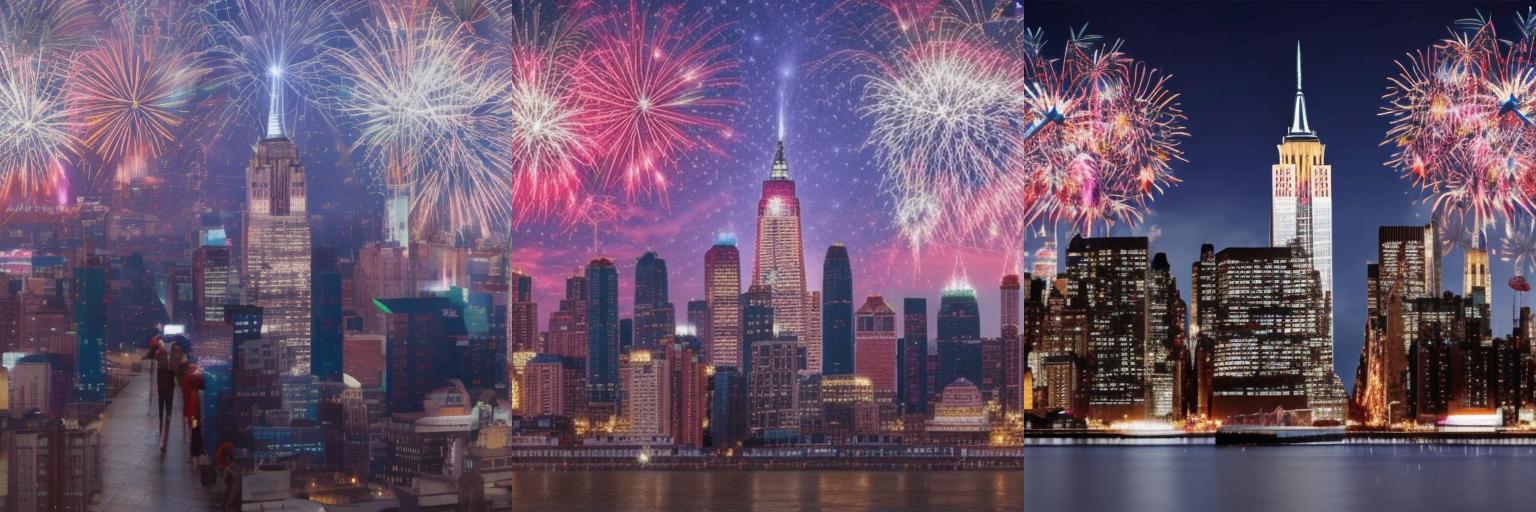}
        \caption{New York Skyline with `Deep Learning' written with fireworks on the sky.}
    \end{subfigure}
    \begin{subfigure}[b]{0.495\textwidth}
        \centering
        \includegraphics[width=\textwidth]{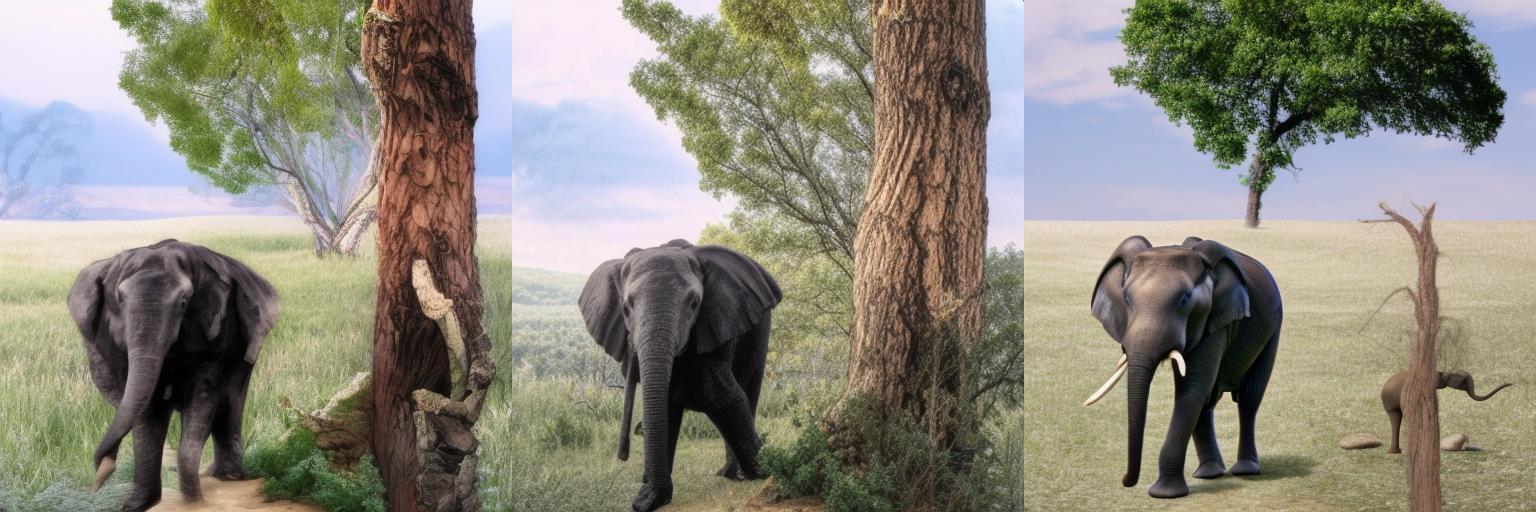}
        \caption{An elephant is behind a tree. You can see the trunk on one side and the back legs on the other.}
    \end{subfigure}
    \begin{subfigure}[b]{0.495\textwidth}
        \centering
        \includegraphics[width=\textwidth]{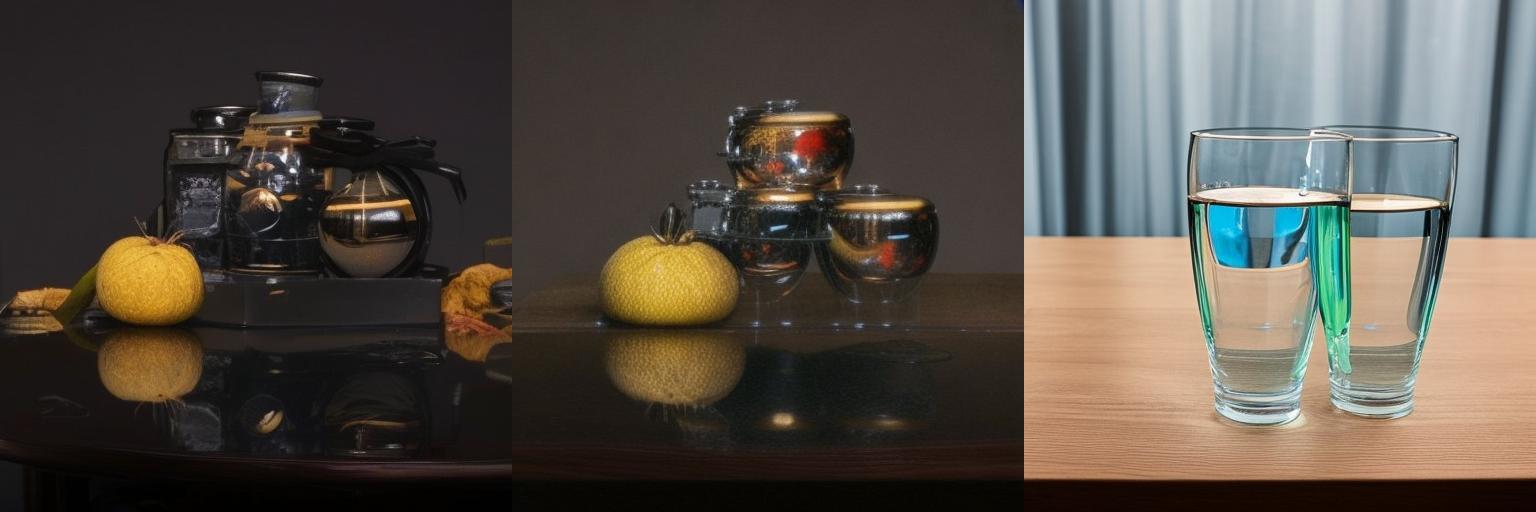}
        \caption{A couple of glasses are sitting on a table.}
    \end{subfigure}
    \begin{subfigure}[b]{0.495\textwidth}
        \centering
        \includegraphics[width=\textwidth]{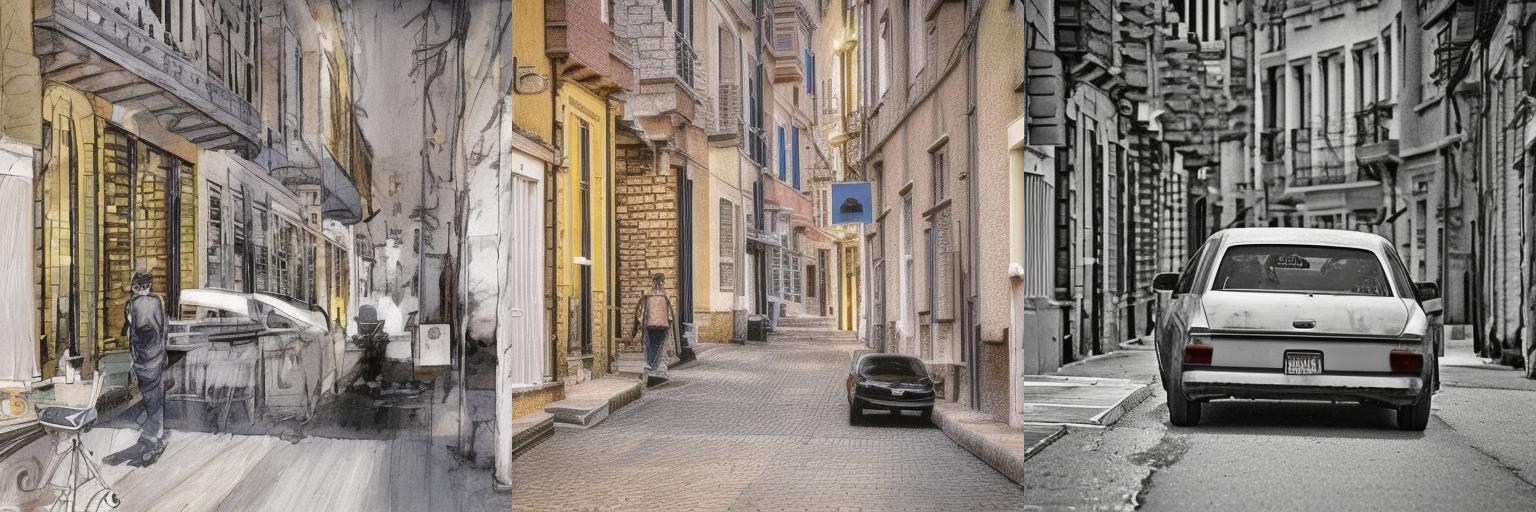}
        \caption{One car on the street.}
    \end{subfigure}
    \begin{subfigure}[b]{0.495\textwidth}
        \centering
        \includegraphics[width=\textwidth]{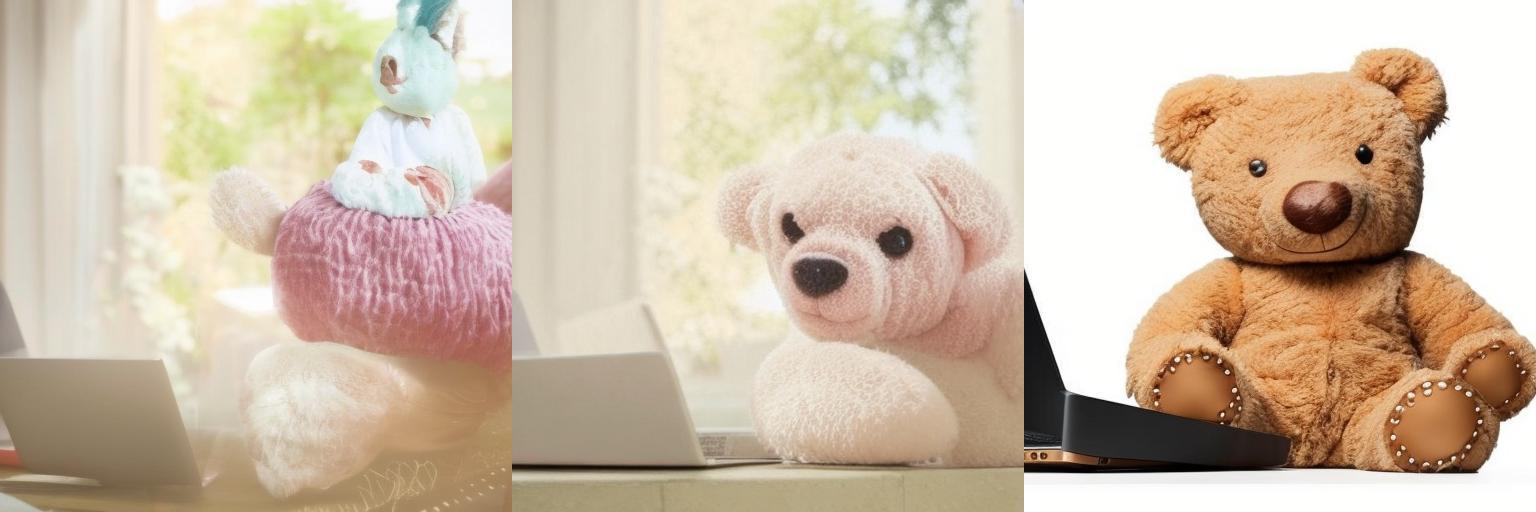}
        \caption{A laptop on top of a teddy bear.}
    \end{subfigure}
    \begin{subfigure}[b]{0.495\textwidth}
        \centering
        \includegraphics[width=\textwidth]{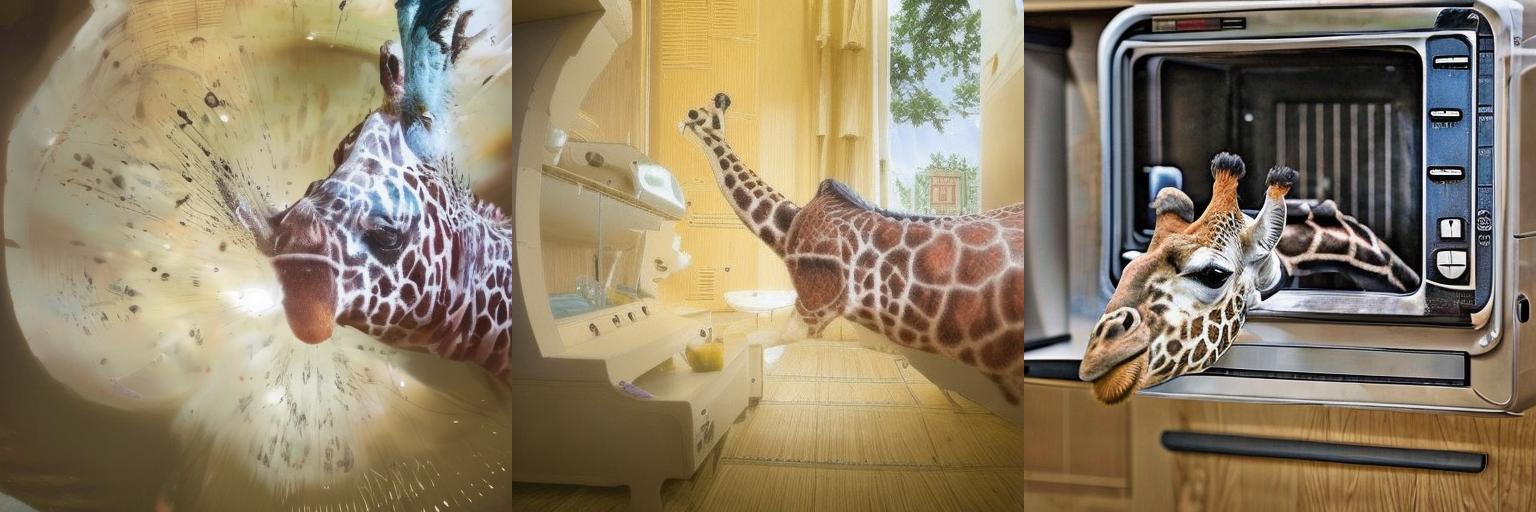}
        \caption{A giraffe underneath a microwave.}
    \end{subfigure}
    \begin{subfigure}[b]{0.495\textwidth}
        \centering
        \includegraphics[width=\textwidth]{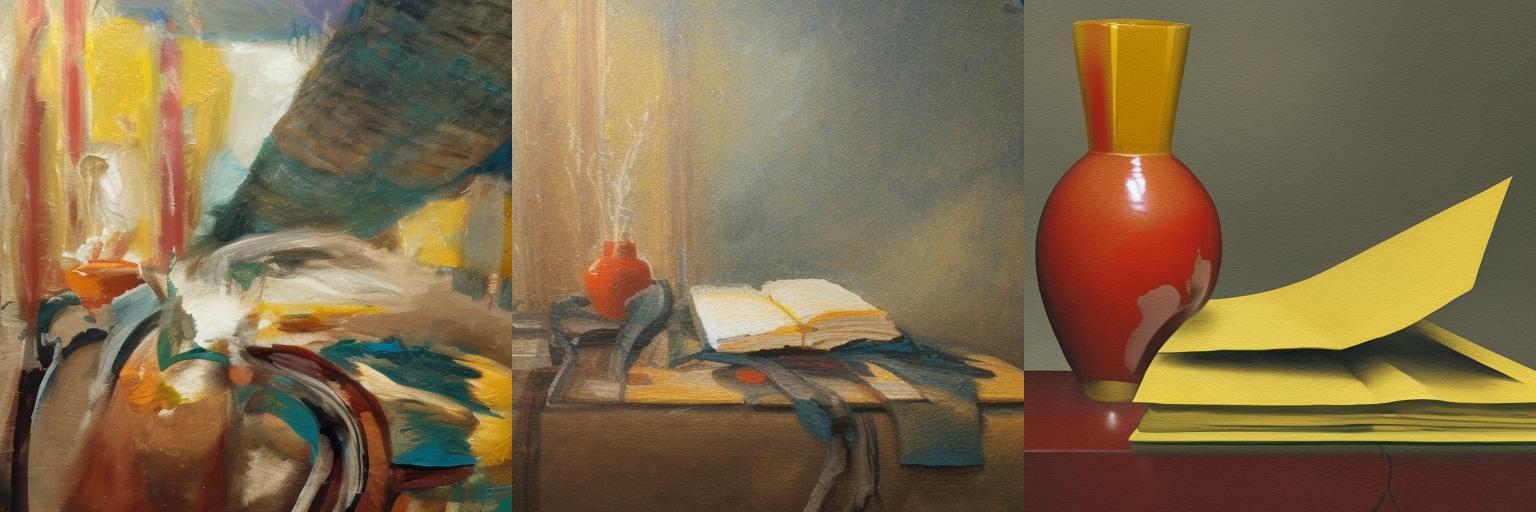}
        \caption{A yellow book and a red vase.}
    \end{subfigure}
    \begin{subfigure}[b]{0.495\textwidth}
        \centering
        \includegraphics[width=\textwidth]{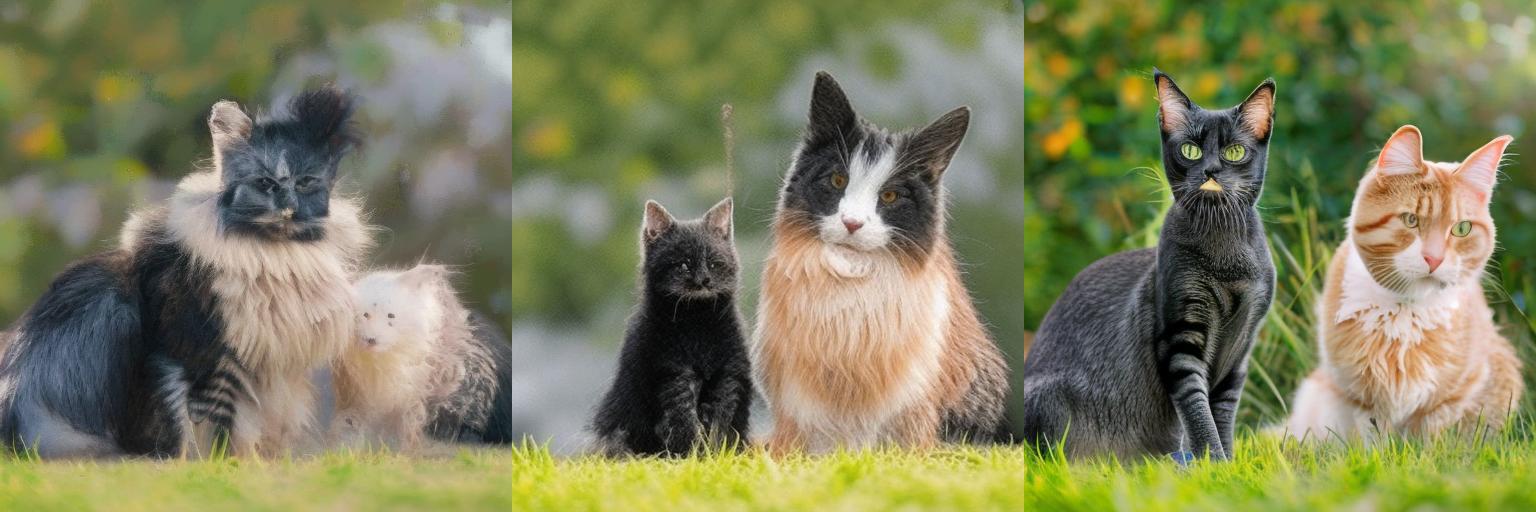}
        \caption{One cat and one dog sitting on the grass.}
    \end{subfigure}
    \begin{subfigure}[b]{0.495\textwidth}
        \centering
        \includegraphics[width=\textwidth]{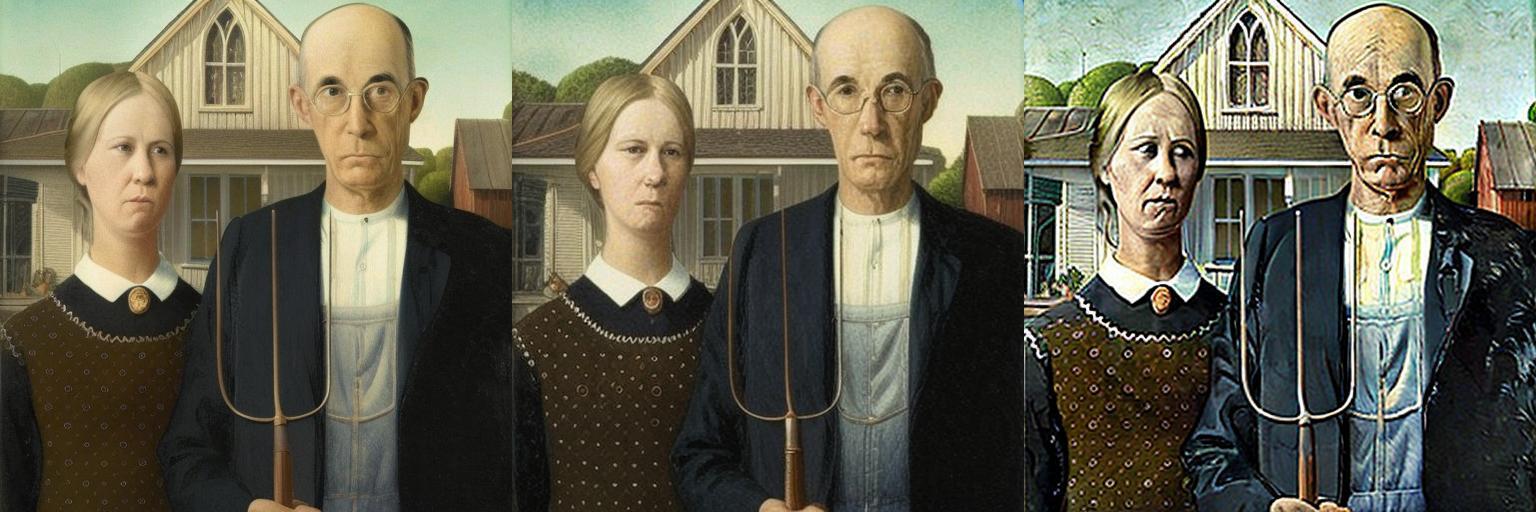}
        \caption{A painting by Grant Wood of an astronaut couple, american gothic style.}
        \label{fig:evaluation-qualitative-painting}
    \end{subfigure}
    \begin{subfigure}[b]{0.495\textwidth}
        \centering
        \includegraphics[width=\textwidth]{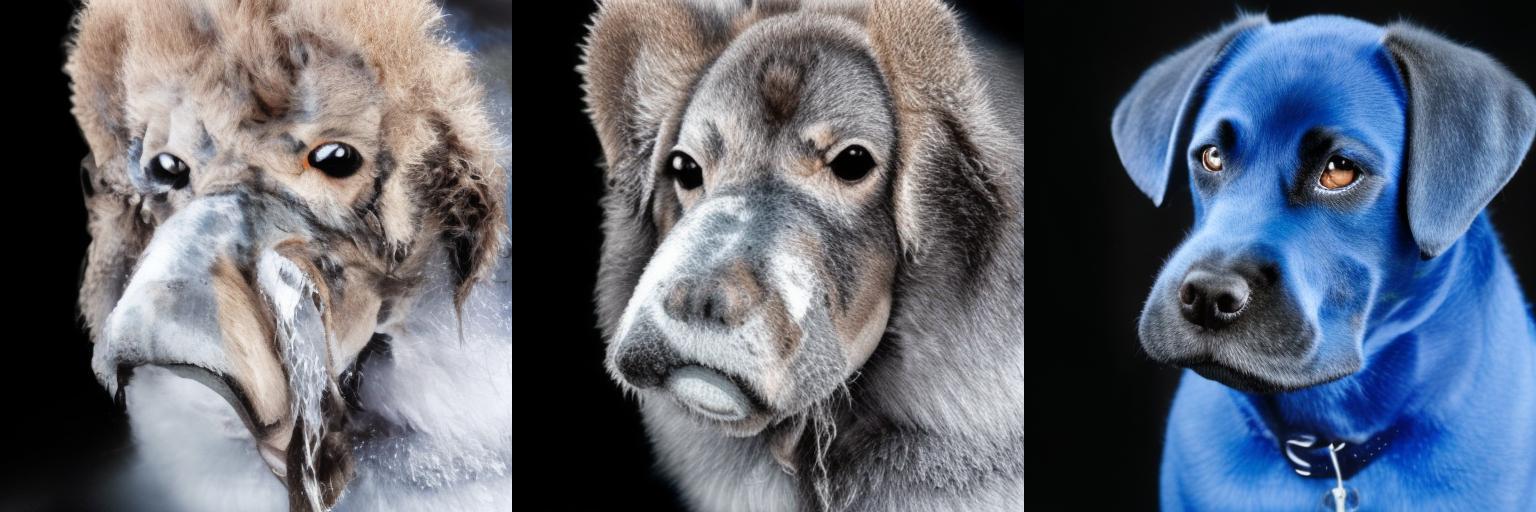}
        \caption{A blue colored dog.}
        \label{fig:evaluation-qualitative-dog}
    \end{subfigure}
    \begin{subfigure}[b]{0.495\textwidth}
        \centering
        \includegraphics[width=\textwidth]{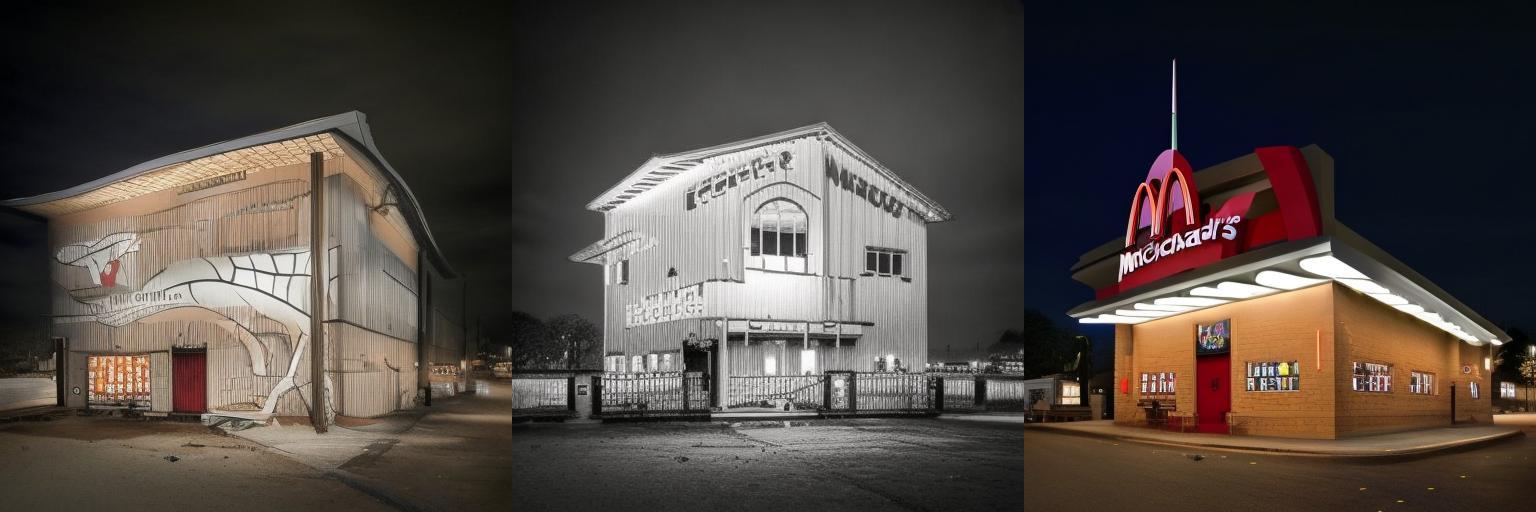}
        \caption{McDonalds Church.}
    \end{subfigure}
    \begin{subfigure}[b]{0.495\textwidth}
        \centering
        \includegraphics[width=\textwidth]{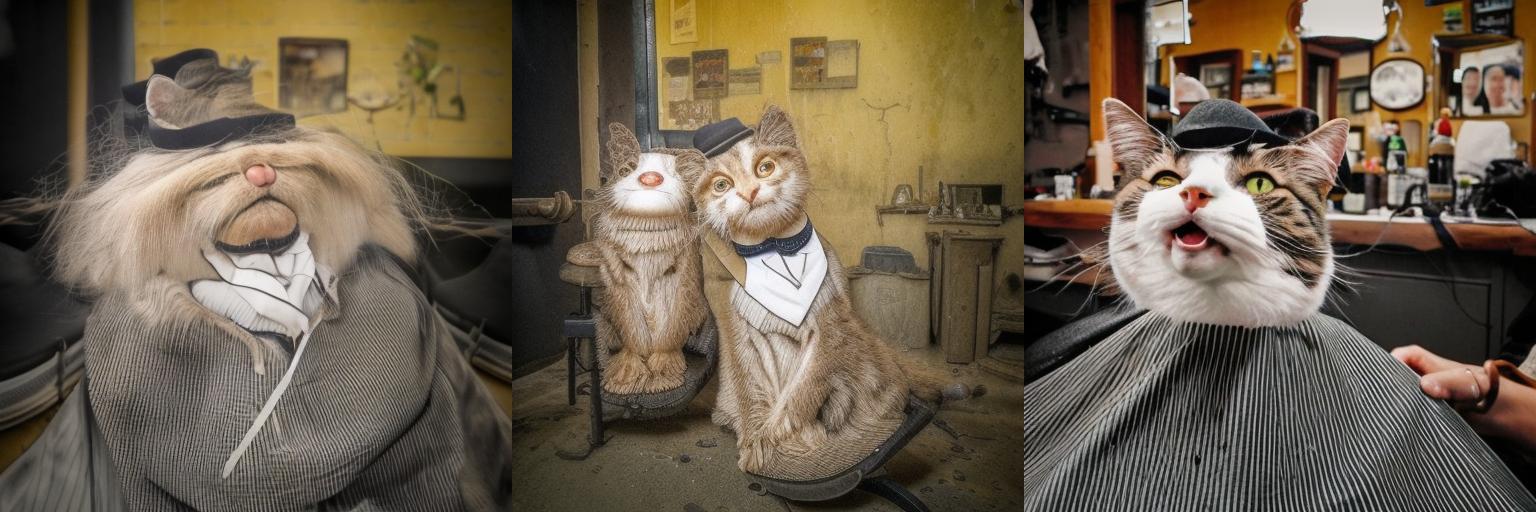}
        \caption{Photo of a cat singing in a barbershop quartet.}
    \end{subfigure}
    \vspace{-5pt}
    \caption{Text-to-image generation on DrawBench prompts \cite{saharia2022photorealistic}. Our self-perceptual objective improves sample quality over the MSE objective while largely maintaining the image content and layout. Classifier-free guidance has the additional effect of enhancing text alignment by sacrificing sample diversity. Images are generated with DDIM 50 NFEs. More analysis in \cref{sec:evaluation-qualitative}.}
    \label{fig:qualitative}
\end{figure*}

\begin{figure*}[t]
    \ContinuedFloat
    \centering
    \captionsetup{justification=raggedright,singlelinecheck=false}
    \small
    \begin{tabularx}{\textwidth}{|X|X|X|X|X|X}
        MSE & Self-Perceptual & MSE + CFG & MSE & Self-Perceptual & MSE + CFG
    \end{tabularx}
    \begin{subfigure}[b]{0.495\textwidth}
        \centering
        \includegraphics[width=\textwidth]{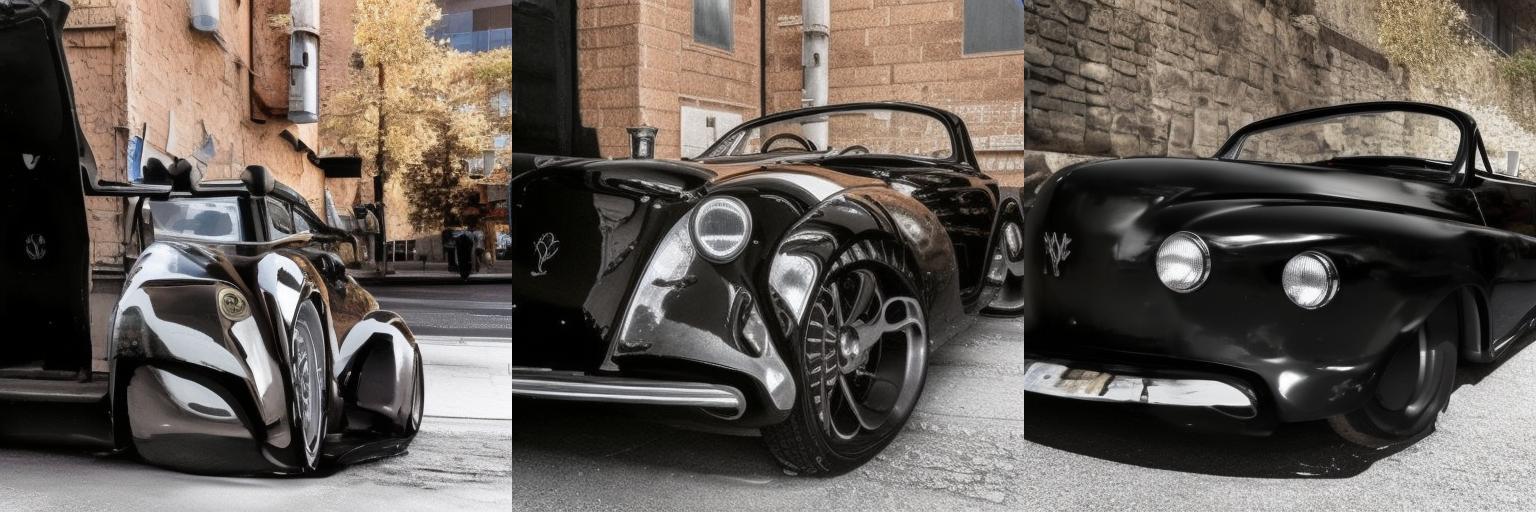}
        \caption{A black colored car.}
    \end{subfigure}
    \begin{subfigure}[b]{0.495\textwidth}
        \centering
        \includegraphics[width=\textwidth]{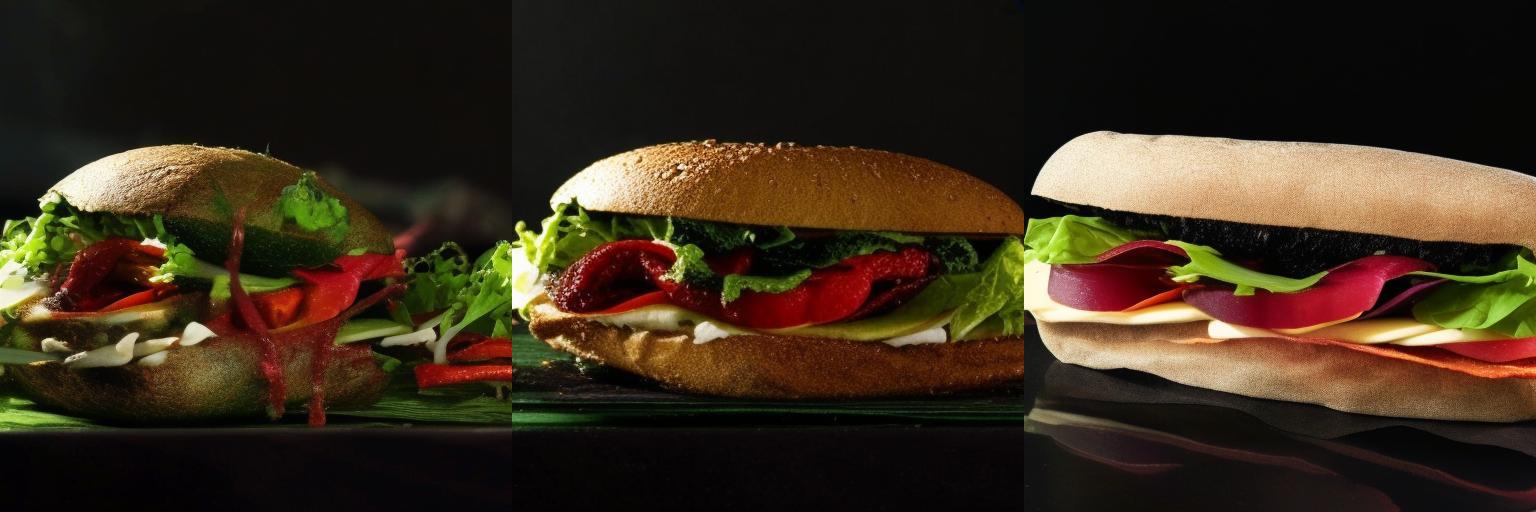}
        \caption{A black colored sandwich.}
    \end{subfigure}
    \begin{subfigure}[b]{0.495\textwidth}
        \centering
        \includegraphics[width=\textwidth]{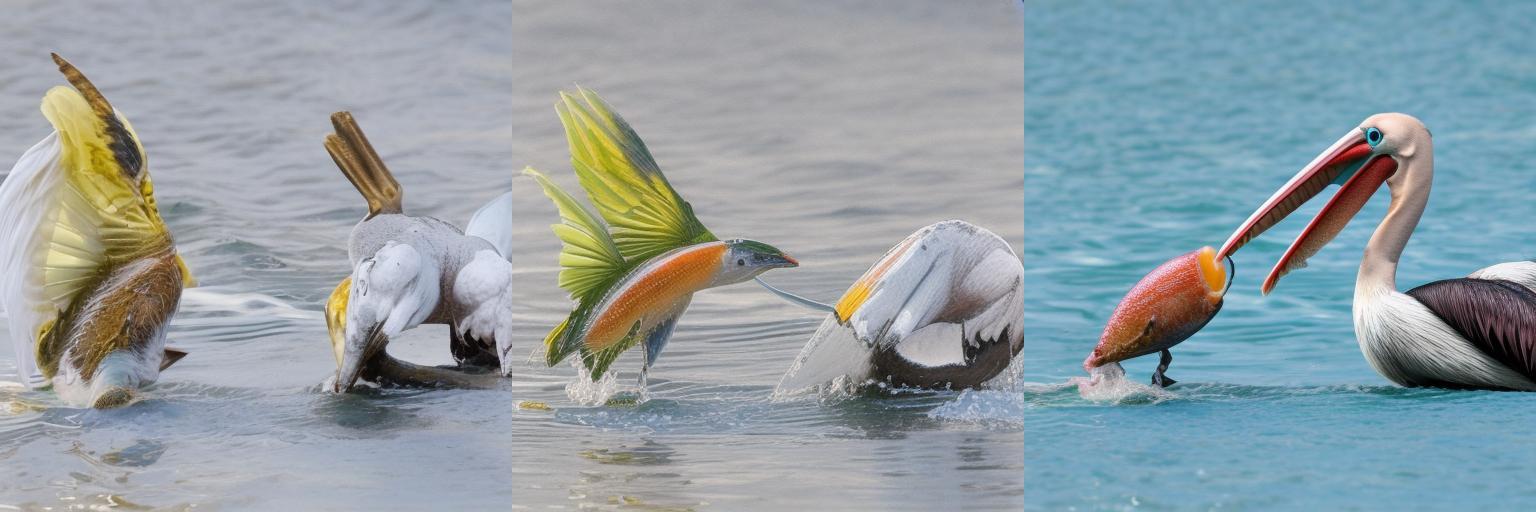}
        \caption{A fish eating a pelican.}
    \end{subfigure}
    \begin{subfigure}[b]{0.495\textwidth}
        \centering
        \includegraphics[width=\textwidth]{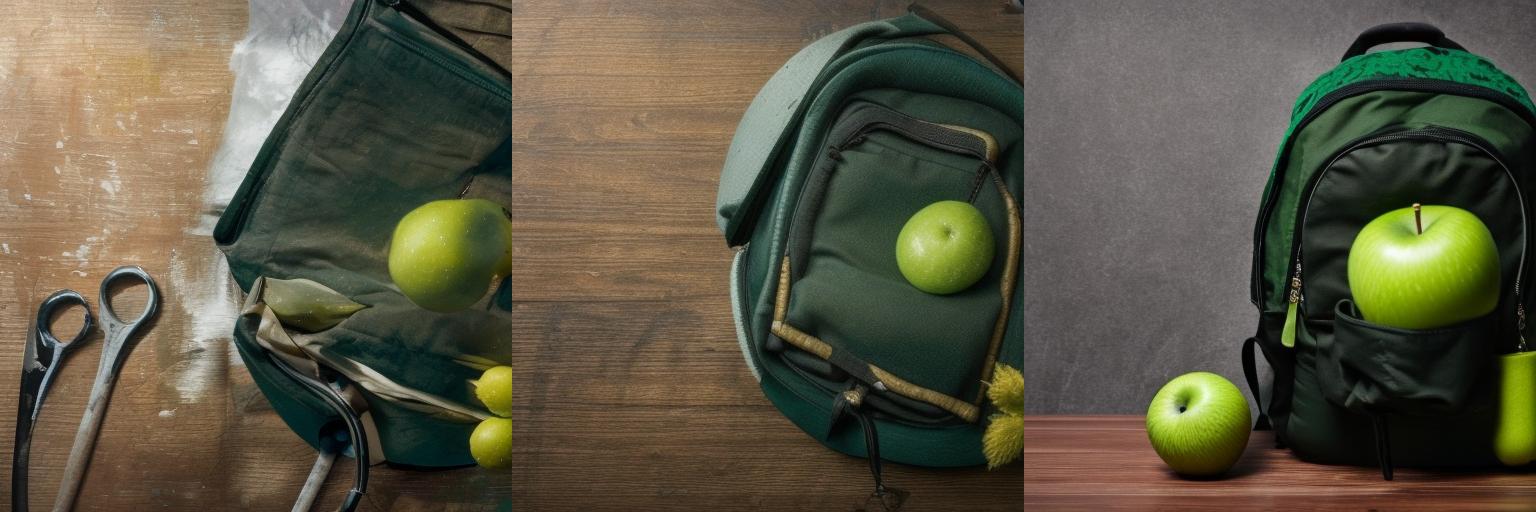}
        \caption{A green apple and a black backpack.}
    \end{subfigure}
    \begin{subfigure}[b]{0.495\textwidth}
        \centering
        \includegraphics[width=\textwidth]{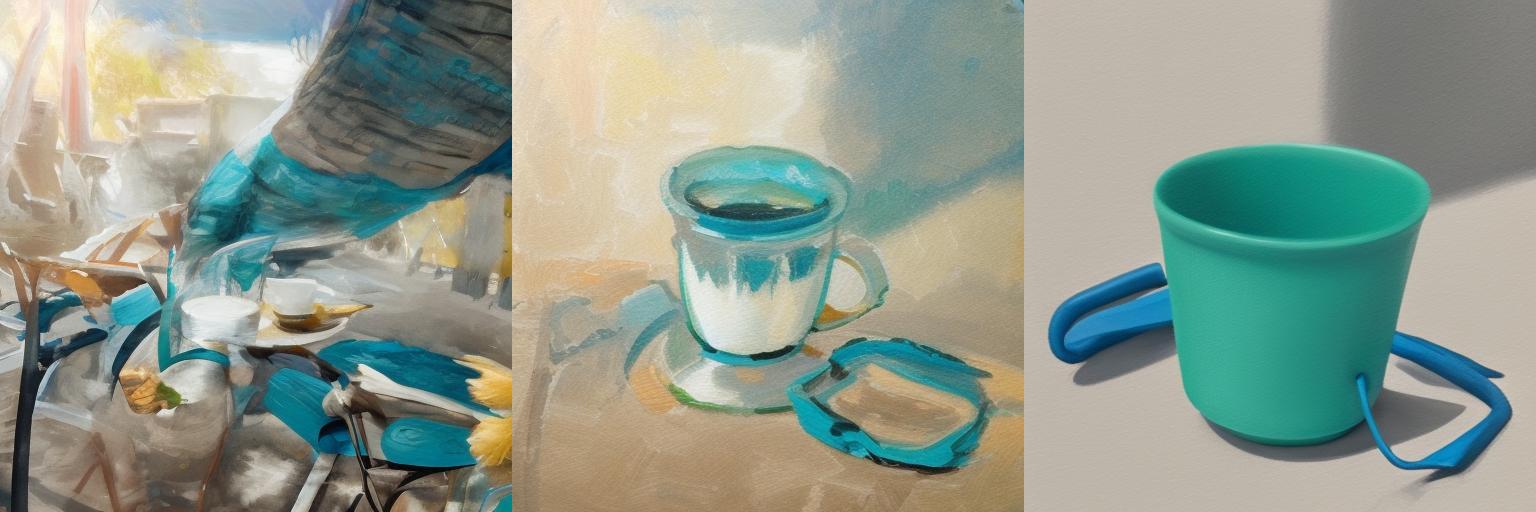}
        \caption{A green cup and a blue cell phone.}
    \end{subfigure}
    \begin{subfigure}[b]{0.495\textwidth}
        \centering
        \includegraphics[width=\textwidth]{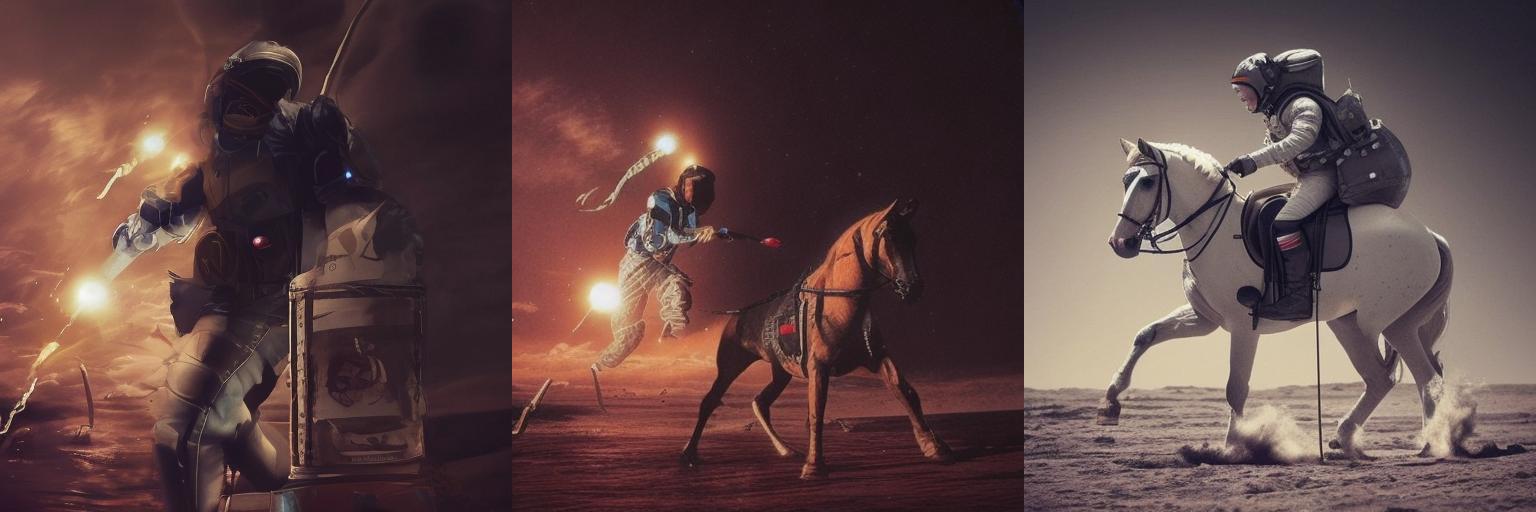}
        \caption{A horse riding an astronaut.}
    \end{subfigure}
    \begin{subfigure}[b]{0.495\textwidth}
        \centering
        \includegraphics[width=\textwidth]{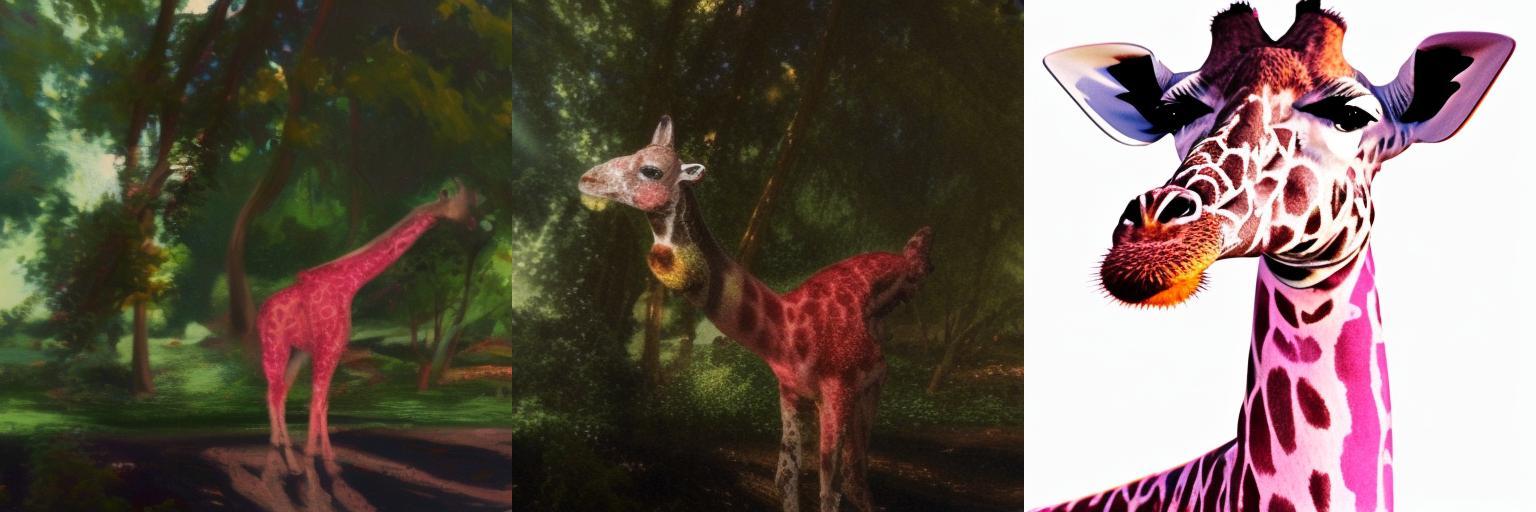}
        \caption{A pink colored giraffe.}
    \end{subfigure}
    \begin{subfigure}[b]{0.495\textwidth}
        \centering
        \includegraphics[width=\textwidth]{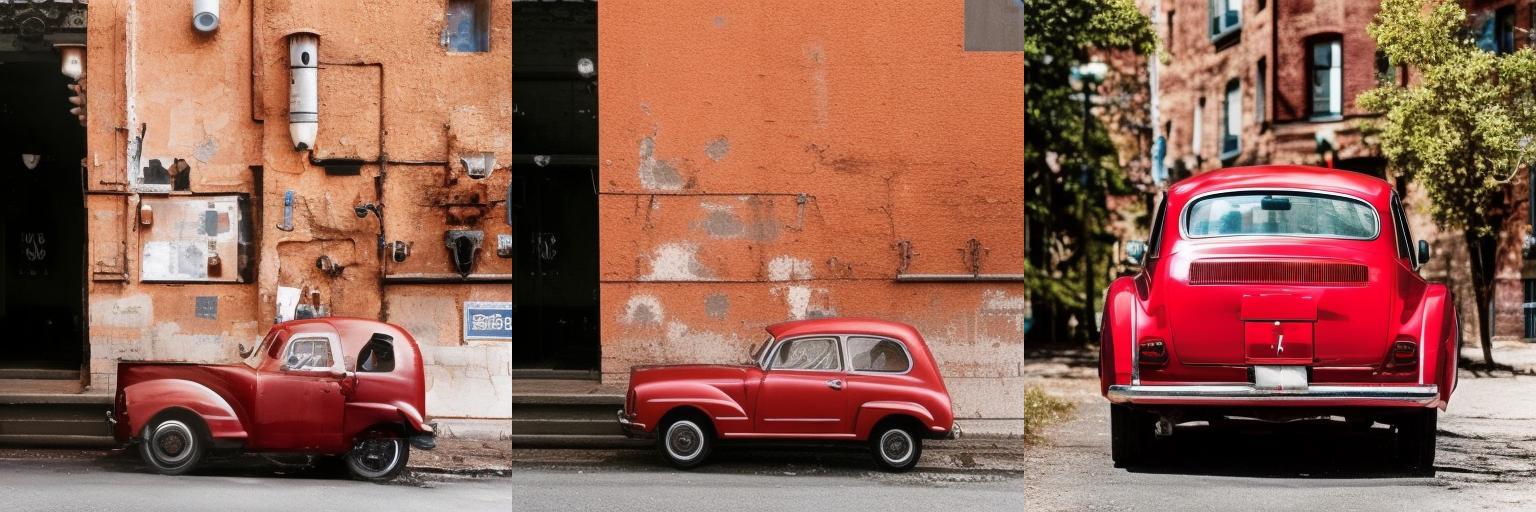}
        \caption{A red colored car.}
    \end{subfigure}
    \begin{subfigure}[b]{0.495\textwidth}
        \centering
        \includegraphics[width=\textwidth]{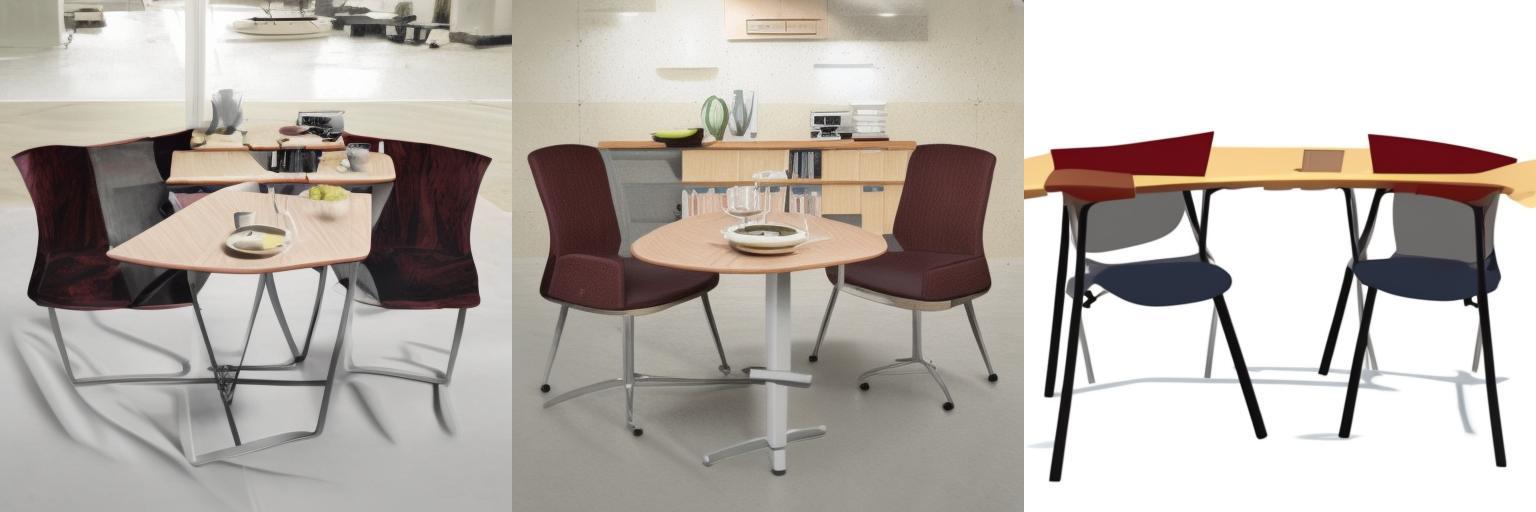}
        \caption{A separate seat for one person, typically with a back and four legs.}
    \end{subfigure}
    \begin{subfigure}[b]{0.495\textwidth}
        \centering
        \includegraphics[width=\textwidth]{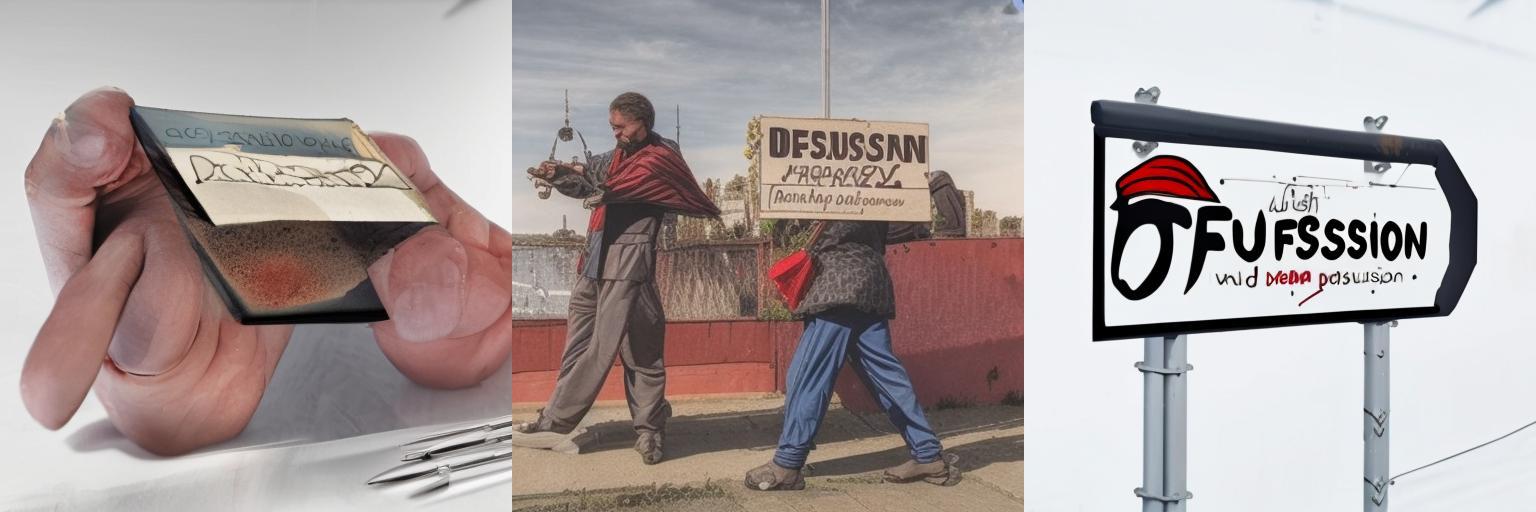}
        \caption{A sign that says `Diffusion'.}
    \end{subfigure}
    \begin{subfigure}[b]{0.495\textwidth}
        \centering
        \includegraphics[width=\textwidth]{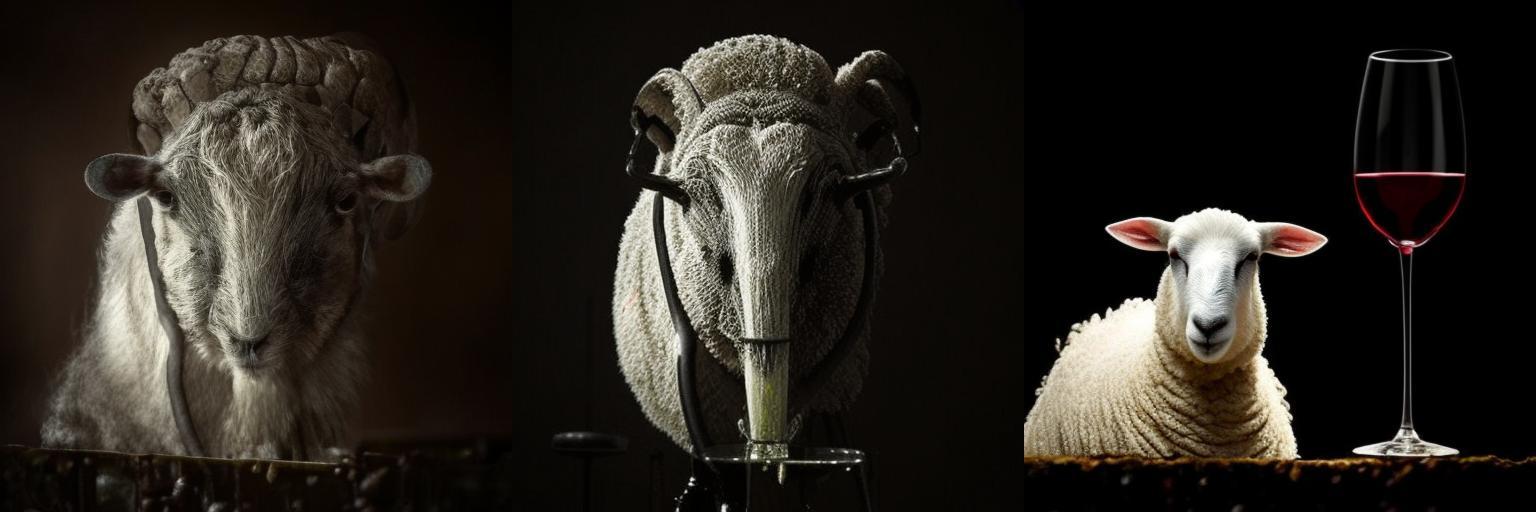}
        \caption{A sheep to the right of a wine glass.}
    \end{subfigure}
    \begin{subfigure}[b]{0.495\textwidth}
        \centering
        \includegraphics[width=\textwidth]{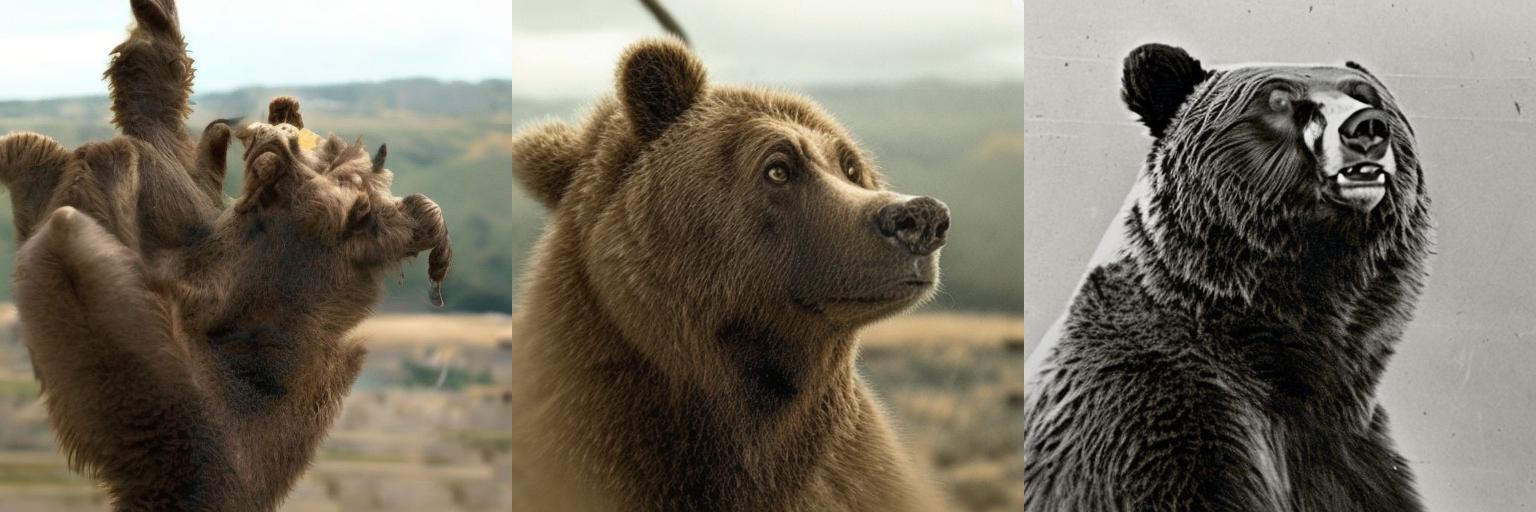}
        \caption{A photo of a confused grizzly bear in calculus class.}
    \end{subfigure}
    \caption{Text-to-image generation on DrawBench prompts \cite{saharia2022photorealistic}. Our self-perceptual objective improves sample quality over the vanilla MSE objective while largely maintaining the image content and layout. Classifier-free guidance has the additional effect of enhancing text alignment by sacrificing sample diversity. Images are generated with DDIM 50 NFEs. More analysis in \cref{sec:evaluation-qualitative}.}
\end{figure*}

\subsection{Quantitative}
\label{sec:evaluation-quantitative}

\Cref{tab:quantitative} shows the quantitative evaluation for conditional generation. We follow the convention to calculate Fréchet Inception Distance (FID) \cite{heusel2018gans,parmar2022aliased} and Inception Score (IS) \cite{salimans2016improved}. We select the first 10k samples from the COCO 2014 validation dataset \cite{lin2015microsoft} and use our models to generate images of the corresponding captions.

Our self-perceptual objective has significantly improved FID/IS over the vanilla MSE objective. This aligns with the improvement observed in our qualitative comparison and validates our analysis. Additionally, we show comparisons with CFG Our SP objective is still weaker compared to CFG. Specifically, we are close to CFG in FID, but CFG+Rescale \cite{lin2023common} is still better in both metrics. However, we emphasize that our main comparison target is the MSE baseline for guidance-less generation. Our research focus is to demonstrate the effect of the loss objective instead of to claim the new state-of-the-art. We discuss limitations and future improvements in \cref{sec:limitation}.

\begin{table}[h]
    \centering
    \setlength\tabcolsep{2.5pt}
    \begin{tabularx}{\linewidth}{Xcccc|cc}
        \toprule
        Loss & CFG & Rescale & Steps & NFE & FID $\downarrow$ & IS $\uparrow$ \\
        \midrule
        Ground truth & & & & & 00.00 & 35.28 \\
        \midrule
        \multirow{2}{*}{$\mathcal{L}_{mse}$} & & & 25 & 25 & 32.68 & 22.20 \\
         & & & 50 & 50 & \textbf{29.63} & \textbf{22.86} \\
        \midrule
        \multirow{2}{*}{$\mathcal{L}_{sp}$} & & &  25 & 25 & 25.89 & 27.76 \\
         & & & 50 & 50 & \textbf{24.42} & \textbf{28.07} \\
        \midrule
        \multirow{2}{*}{$\mathcal{L}_{mse}$} & 7.5 & & 25 & 50 & 24.41 & 32.10 \\
         & 7.5 & 0.7  & 25 & 50 & \textbf{18.67} & \textbf{34.17} \\
        \bottomrule
    \end{tabularx}
    \vspace{-0.5em}
    \caption{Conditional generation. Quantitative evaluation on MSCOCO 10K validation dataset. Our self-perceptual (SP) objective improves FID and IS metrics over MSE objective but has not surpassed classifier-free guidance \cite{ho2022classifierfree} with rescale \cite{lin2023common}. Since classifier-free guidance with 25 steps incurs 50 NFEs (number of function evaluations), we show both 25-step and 50-step metrics.}
    \label{tab:quantitative}
\end{table}

\begin{table}[h!]
    \centering
    \setlength\tabcolsep{2.5pt}
    \begin{tabularx}{\linewidth}{Xcc}
        \toprule
        Loss & FID $\downarrow$ & IS $\uparrow$ \\
        \midrule
        $\mathcal{L}_{mse}$ & 62.32 & 11.18 \\
        $\mathcal{L}_{sp}$ & \textbf{59.12} & \textbf{12.04} \\
        \bottomrule
    \end{tabularx}
    \vspace{-0.5em}
    \caption{Unconditional generation metrics. The self-perceptual objective improves FID and IS over the MSE objective.}
    \label{tab:quantitative-uncond}
    \vspace{-1em}
\end{table}

\Cref{tab:quantitative-uncond} shows that our approach also improves FID/IS for unconditional generation.

\section{Ablation Study}

In this section, we evaluate the individual hyperparameters for our self-perceptual objective. All metrics are calculated on the same MSCOCO 10k validation samples as in \cref{sec:evaluation-quantitative} and use 25 steps of DDIM inference.

\subsection{Layer $l$: Only Midblock Layer is Better}
\label{sec:ablation-layer}

We compare the effect of computing loss on features from different layers $l$. We find that only using the features from the midblock layer yields better results, as shown in \cref{tab:ablation-layer}.

\begin{table}[h]
    \centering
    \setlength\tabcolsep{2.5pt}
    \begin{tabularx}{\linewidth}{Xccc}
        \toprule
        Layer & FID $\downarrow$ & IS $\uparrow$ \\
        \midrule
        All Encoder Layers & 26.64 & 26.89 \\
        All Decoder Layers & 42.42 & 19.98 \\
        All Encoder Layers + Midblock Layer & 26.96 & 27.24 \\
        Only Midblock Layer & \textbf{25.89} & \textbf{27.76} & \checkmark  \\
        \bottomrule
    \end{tabularx}
    \vspace{-0.5em}
    \caption{Comparing computing perceptual loss on different layers. We find that only computing loss on the midblock hidden features yields better results.}
    \label{tab:ablation-layer}
    \vspace{-1em}
\end{table}

\subsection{Timestep $t'$: Uniform Sampling is Better}
\label{sec:ablation-timestep}

We compare the effect of selecting timestep $t'$ for the perceptual network. First, notice that $t'=t$ is invalid because $\hat{x}_t$ always equals $x_t$, which makes the input to the perceptual network identical and prevents meaningful loss. We compare three different choices for $t'$ in \cref{tab:ablation-timestep} and show that uniform sampling of $t'$ yields good results.

\begin{table}[h!]
    \centering
    \setlength\tabcolsep{2.5pt}
    \begin{tabularx}{\linewidth}{Xccc}
        \toprule
        Timestep ($t'$ clamped to $[1, T]$) & FID $\downarrow$ & IS $\uparrow$ \\
        \midrule
        $t'=t \pm 40$ (1000 / 25 steps = 40) & 27.24 & 23.31 \\
        $t'\sim \mathcal{N}(t, 100)$ & \textbf{24.54} & 25.42 \\
        $t'\sim \mathcal{U}(1, T)$ & 25.89 & \textbf{27.76} & \checkmark  \\
        \bottomrule
    \end{tabularx}
    \vspace{-0.5em}
    \caption{Comparing the choice of timestep $t'$. We find that simply uniformly sampling $t'$ can yield reasonably good results.}
    \label{tab:ablation-timestep}
    \vspace{-1em}
\end{table}

\subsection{Feature Distance Function: Not Influential}

We compare using different distance functions on the hidden features. \Cref{tab:ablation-distance-function} shows that MSE and MAE yield similar results, so we stick to MSE.

\begin{table}[h!]
    \centering
    \setlength\tabcolsep{2.5pt}
    \begin{tabularx}{\linewidth}{lXccc}
        \toprule
        Distance & & FID $\downarrow$ & IS $\uparrow$ \\
        \midrule
        Mean Absolute Distance & ($\| \cdot \|_1$) & \textbf{25.28} & 27.41 \\
        Mean Squared Distance & ($\| \cdot \|_2^2$) & 25.89 & \textbf{27.76} & \checkmark \\
        \bottomrule
    \end{tabularx}
    \vspace{-0.5em}
    \caption{Comparing the choice of distance function. We find that mean squared distance and mean absolute distance have similar results, so we stick to mean squared distance.}
    \label{tab:ablation-distance-function}
    \vspace{-1em}
\end{table}

% \subsection{Other Possible Formulations}

% We experiment with an alternative formulation, which combines predicted $\hat{x}_{t'},\hat{\epsilon}_{t'}$ separately with ground-truth $x_{t'},\epsilon_{t'}$. This formulation allows gradient feedback at $t'=t$:
% \begin{align}
%     x_{t'} &= \mathbf{forward}(x_0, \epsilon, t'),\\
%     \hat{x}_{t'}^{x} &= \mathbf{forward}(\hat{x}_0, \epsilon, t'),\\
%     \hat{x}_{t'}^{\epsilon} &= \mathbf{forward}(x_0, \hat\epsilon, t'),
% \end{align}
% \begin{equation}
% \begin{aligned}
%     \mathcal{L}_{sp2}  = & \| f^l_*(\hat{x}_{t'}^x, t', c) - f^l_*(x_{t'}, t', c) \|_2^2 \\
%      + & \| f^l_*(\hat{x}_{t'}^\epsilon, t', c) - f^l_*(x_{t'}, t', c) \|_2^2.
% \end{aligned}
% \end{equation}

% \Cref{tab:ablation-formulation} shows that the alternative formulation yields worse performance.

% \begin{table}[h!]
%     \centering
%     \setlength\tabcolsep{2.5pt}
%     \begin{tabularx}{\linewidth}{Xccc}
%         \toprule
%         Formulation & FID $\downarrow$ & IS $\uparrow$ \\
%         \midrule
%         $\mathcal{L}_{sp}$ & \textbf{25.89} & \textbf{27.76} & \checkmark \\
%         $\mathcal{L}_{sp2}$ & 29.83 & 24.54 \\
%         \bottomrule
%     \end{tabularx}
%     \vspace{-0.5em}
%     \caption{Comparing different formulations. We find that merged formulation yields the best results.}
%     \label{tab:ablation-formulation}
%     \vspace{-1em}
% \end{table}

\subsection{Repeat Perceptual Network}

We experiment using the network trained with self-perceptual objective as the perceptual metric network $f^l_*$ and repeat the training process. \Cref{tab:ablation-repeat} shows that repeating the self-perceptual training results in worse performance. This is why we decide to just freeze the MSE model instead of using an exponential moving average (EMA) for the perceptual network.

\begin{table}[h!]
    \centering
    \setlength\tabcolsep{2.5pt}
    \begin{tabularx}{\linewidth}{Xccc}
        \toprule
        Formulation & FID $\downarrow$ & IS $\uparrow$ \\
        \midrule
        MSE model as perceptual network & \textbf{25.89} & \textbf{27.76} & \checkmark \\
        SP model as perceptual network & 26.61 & 26.41 \\
        \bottomrule
    \end{tabularx}
    \vspace{-0.5em}
    \caption{Repeating the self-perceptual process yields worse performance.}
    \label{tab:ablation-repeat}
    \vspace{-1em}
\end{table}

\subsection{Combine with Classifier-Free Guidance}

\begin{table}[h]
    \centering
    \setlength\tabcolsep{2.5pt}
    \begin{tabularx}{\linewidth}{Xcc|cc}
        \toprule
        Loss & CFG & Rescale & FID $\downarrow$ & IS $\uparrow$ \\
        \midrule
        $\mathcal{L}_{mse}$ & 7.5 & 0.7  & \textbf{18.67} & \textbf{34.17} \\
        \midrule
        \multirow{5}{*}{$\mathcal{L}_{sp}$} & & & 25.89 & 27.76 \\
         & 2.0 & 0.7 & 21.19 & 32.22 \\
         & 3.0 & 0.7 & \textbf{20.65} & \textbf{33.49} \\
         & 4.0 & 0.7 & 20.67 & 33.34 \\
         & 7.5 & 0.7 & 23.49 & 31.64 \\
        \bottomrule
    \end{tabularx}
    \vspace{-0.8em}
    \caption{Combining our self-perceptual objective with classifier-free guidance does improve sample quality but does not surpass the MSE objective with classifier-free guidance.}
    \label{tab:ablation-cfg}
    \vspace{-1.2em}
\end{table}

We experiment with applying classifier-free guidance on the model trained with our self-perceptual objective. \Cref{tab:ablation-cfg} shows that classifier-free guidance indeed can improve sample quality further on the self-perceptual model but it does not surpass classifier-free guidance applied on the MSE model. We find artifacts as described in \cref{sec:ablation-behavior}.

\subsection{Artifacts Caused by the Perceptual Network}
\label{sec:ablation-behavior}

\begin{figure}[h]
    \centering
    \captionsetup{justification=raggedright,singlelinecheck=false}
    \footnotesize
    \setlength\tabcolsep{0pt}
    \newcolumntype{Y}{>{\centering\arraybackslash}X}
    \begin{tabularx}{\linewidth}{|Y|Y|Y|Y|Y|Y|Y|Y|Y|Y|Y}
        1000 & 900 & 800 & 700 & 600 & 500 & 400 & 300 & 200 & 100 & Final
    \end{tabularx}
    \begin{subfigure}[b]{\linewidth}
        \centering
        \includegraphics[width=\textwidth]{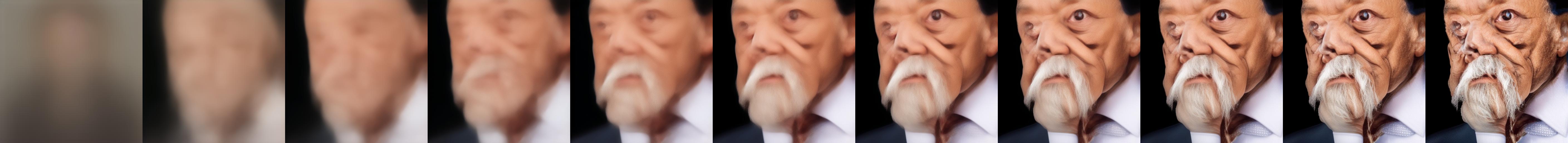}
        \caption{MSE}
    \end{subfigure}
    \begin{subfigure}[b]{\linewidth}
        \centering
        \includegraphics[width=\linewidth]{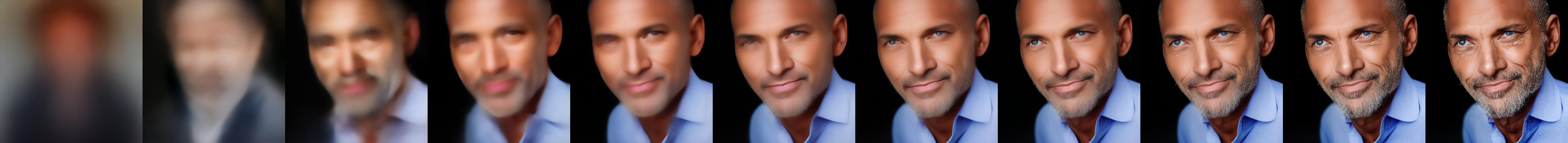}
        \caption{MSE + CFG}
    \end{subfigure}
    \begin{subfigure}[b]{\linewidth}
        \centering
        \includegraphics[width=\linewidth]{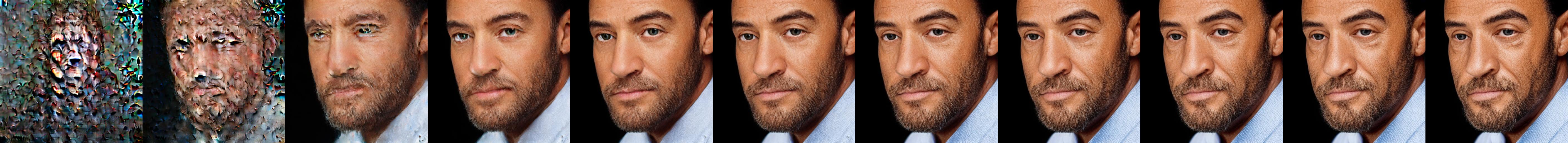}
        \caption{Self-Perceptual}
    \end{subfigure}
    \vspace{-2em}
    \caption{Model prediction at each step converted to the $\hat{x}_0$ space.}
    \label{fig:ablation-behavior}
    \vspace{-1em}
\end{figure}

In \cref{fig:ablation-behavior}, we visualize the model prediction by converting to $\hat{x}_0$ at every inference step. We see grid-like pattern artifacts resulting from the perceptual network using convolution with kernel size 3 and stride 2 for downsampling \cite{odena2016deconvolution}. This can be an area for minor future improvement.

\subsection{Limitations and Future Works}
\label{sec:limitation}
In this work, we compute distance on the features of a frozen network. Although this exhibits better perceptual alignment than MSE, it is still not ideal. Our proposed SP objective is only meant to validate the effect of the loss objective and is not proposed as a final solution. We believe the loss objective should ultimately be learned. This draws a connection to adversarial training where the discriminator is a learnable perceptual network in the loop. We leave the exploration to future work.

\section{Conclusion}

Our paper elucidates that the loss objective has an important role in shaping the learned distribution of diffusion models. We show that the MSE loss causes the diffusion models to generate poor samples without guidance and demonstrate the effectiveness of perceptual loss in improving sample quality. We hope our work paves the way for more future explorations on diffusion training objectives.

\section*{Acknowledgment}
We thank Lu Jiang and Ceyuan Yang for reviewing the manuscript and providing valuable feedback.

{
    \small
    \bibliographystyle{ieeenat_fullname}
    \bibliography{main}

\begin{thebibliography}{57}
\providecommand{\natexlab}[1]{#1}
\providecommand{\url}[1]{\texttt{#1}}
\expandafter\ifx\csname urlstyle\endcsname\relax
  \providecommand{\doi}[1]{doi: #1}\else
  \providecommand{\doi}{doi: \begingroup \urlstyle{rm}\Url}\fi

\bibitem[Blattmann et~al.(2023{\natexlab{a}})Blattmann, Dockhorn, Kulal, Mendelevitch, Kilian, Lorenz, Levi, English, Voleti, Letts, Jampani, and Rombach]{blattmann2023stable}
Andreas Blattmann, Tim Dockhorn, Sumith Kulal, Daniel Mendelevitch, Maciej Kilian, Dominik Lorenz, Yam Levi, Zion English, Vikram Voleti, Adam Letts, Varun Jampani, and Robin Rombach.
\newblock Stable video diffusion: Scaling latent video diffusion models to large datasets, 2023{\natexlab{a}}.

\bibitem[Blattmann et~al.(2023{\natexlab{b}})Blattmann, Rombach, Ling, Dockhorn, Kim, Fidler, and Kreis]{blattmann2023align}
A. Blattmann, Robin Rombach, Huan Ling, Tim Dockhorn, Seung~Wook Kim, Sanja Fidler, and Karsten Kreis.
\newblock Align your latents: High-resolution video synthesis with latent diffusion models.
\newblock \emph{2023 IEEE/CVF Conference on Computer Vision and Pattern Recognition (CVPR)}, pages 22563--22575, 2023{\natexlab{b}}.

\bibitem[Chang et~al.(2023)Chang, Shi, Gao, Fu, Xu, Song, Yan, Yang, and Soleymani]{chang2023magicdance}
Di Chang, Yichun Shi, Quankai Gao, Jessica Fu, Hongyi Xu, Guoxian Song, Qing Yan, Xiao Yang, and Mohammad Soleymani.
\newblock Magicdance: Realistic human dance video generation with motions \& facial expressions transfer, 2023.

\bibitem[Chen et~al.(2025)Chen, Jiang, Zheng, Chen, Su, and Zhu]{chen2025visualgenerationguidance}
Huayu Chen, Kai Jiang, Kaiwen Zheng, Jianfei Chen, Hang Su, and Jun Zhu.
\newblock Visual generation without guidance, 2025.

\bibitem[Chen et~al.(2024)Chen, YU, GE, Yao, Xie, Wang, Kwok, Luo, Lu, and Li]{chen2023pixartalpha}
Junsong Chen, Jincheng YU, Chongjian GE, Lewei Yao, Enze Xie, Zhongdao Wang, James Kwok, Ping Luo, Huchuan Lu, and Zhenguo Li.
\newblock Pixart-\${\textbackslash}alpha\$: Fast training of diffusion transformer for photorealistic text-to-image synthesis.
\newblock In \emph{The Twelfth International Conference on Learning Representations}, 2024.

\bibitem[Choi et~al.(2022)Choi, Lee, Shin, Kim, Kim, and Yoon]{Choi2022PerceptionPT}
Jooyoung Choi, Jungbeom Lee, Chaehun Shin, Sungwon Kim, Hyunwoo~J. Kim, and Sung-Hoon Yoon.
\newblock Perception prioritized training of diffusion models.
\newblock \emph{2022 IEEE/CVF Conference on Computer Vision and Pattern Recognition (CVPR)}, pages 11462--11471, 2022.

\bibitem[Dhariwal and Nichol(2021)]{dhariwal2021diffusion}
Prafulla Dhariwal and Alexander~Quinn Nichol.
\newblock Diffusion models beat gans on image synthesis.
\newblock In \emph{Advances in Neural Information Processing Systems 34: Annual Conference on Neural Information Processing Systems 2021, NeurIPS 2021, December 6-14, 2021, virtual}, pages 8780--8794, 2021.

\bibitem[Esser et~al.(2023)Esser, Chiu, Atighehchian, Granskog, and Germanidis]{esser2023structure}
Patrick Esser, Johnathan Chiu, Parmida Atighehchian, Jonathan Granskog, and Anastasis Germanidis.
\newblock Structure and content-guided video synthesis with diffusion models.
\newblock \emph{2023 IEEE/CVF International Conference on Computer Vision (ICCV)}, pages 7312--7322, 2023.

\bibitem[Esser et~al.(2024)Esser, Kulal, Blattmann, Entezari, Müller, Saini, Levi, Lorenz, Sauer, Boesel, Podell, Dockhorn, English, Lacey, Goodwin, Marek, and Rombach]{esser2024scalingrectifiedflowtransformers}
Patrick Esser, Sumith Kulal, Andreas Blattmann, Rahim Entezari, Jonas Müller, Harry Saini, Yam Levi, Dominik Lorenz, Axel Sauer, Frederic Boesel, Dustin Podell, Tim Dockhorn, Zion English, Kyle Lacey, Alex Goodwin, Yannik Marek, and Robin Rombach.
\newblock Scaling rectified flow transformers for high-resolution image synthesis, 2024.

\bibitem[Goodfellow et~al.(2014)Goodfellow, Pouget-Abadie, Mirza, Xu, Warde-Farley, Ozair, Courville, and Bengio]{goodfellow2014generative}
Ian~J. Goodfellow, Jean Pouget-Abadie, Mehdi Mirza, Bing Xu, David Warde-Farley, Sherjil Ozair, Aaron~C. Courville, and Yoshua Bengio.
\newblock Generative adversarial networks.
\newblock \emph{Communications of the ACM}, 63:\penalty0 139 -- 144, 2014.

\bibitem[Guo et~al.(2023)Guo, Xu, Pu, Ni, Wang, Vasu, Song, Huang, and Shi]{Guo2023SmoothDC}
Jiayi Guo, Xingqian Xu, Yifan Pu, Zanlin Ni, Chaofei Wang, Manushree Vasu, Shiji Song, Gao Huang, and Humphrey Shi.
\newblock Smooth diffusion: Crafting smooth latent spaces in diffusion models.
\newblock \emph{2024 IEEE/CVF Conference on Computer Vision and Pattern Recognition (CVPR)}, pages 7548--7558, 2023.

\bibitem[Guo et~al.(2024)Guo, Yang, Rao, Liang, Wang, Qiao, Agrawala, Lin, and Dai]{guo2023animatediff}
Yuwei Guo, Ceyuan Yang, Anyi Rao, Zhengyang Liang, Yaohui Wang, Yu Qiao, Maneesh Agrawala, Dahua Lin, and Bo Dai.
\newblock Animatediff: Animate your personalized text-to-image diffusion models without specific tuning.
\newblock In \emph{The Twelfth International Conference on Learning Representations}, 2024.

\bibitem[Hang et~al.(2023)Hang, Gu, Li, Bao, Chen, Hu, Geng, and Guo]{Hang2023EfficientDT}
Tiankai Hang, Shuyang Gu, Chen Li, Jianmin Bao, Dong Chen, Han Hu, Xin Geng, and Baining Guo.
\newblock Efficient diffusion training via min-snr weighting strategy.
\newblock \emph{2023 IEEE/CVF International Conference on Computer Vision (ICCV)}, pages 7407--7417, 2023.

\bibitem[Heusel et~al.(2017)Heusel, Ramsauer, Unterthiner, Nessler, and Hochreiter]{heusel2018gans}
Martin Heusel, Hubert Ramsauer, Thomas Unterthiner, Bernhard Nessler, and Sepp Hochreiter.
\newblock Gans trained by a two time-scale update rule converge to a local nash equilibrium.
\newblock In \emph{Advances in Neural Information Processing Systems 30: Annual Conference on Neural Information Processing Systems 2017, December 4-9, 2017, Long Beach, CA, {USA}}, pages 6626--6637, 2017.

\bibitem[Ho and Salimans(2021)]{ho2022classifierfree}
Jonathan Ho and Tim Salimans.
\newblock Classifier-free diffusion guidance.
\newblock In \emph{NeurIPS 2021 Workshop on Deep Generative Models and Downstream Applications}, 2021.

\bibitem[Ho et~al.(2020)Ho, Jain, and Abbeel]{ho2020denoising}
Jonathan Ho, Ajay Jain, and Pieter Abbeel.
\newblock Denoising diffusion probabilistic models.
\newblock In \emph{Advances in Neural Information Processing Systems 33: Annual Conference on Neural Information Processing Systems 2020, NeurIPS 2020, December 6-12, 2020, virtual}, 2020.

\bibitem[Ho et~al.(2022)Ho, Chan, Saharia, Whang, Gao, Gritsenko, Kingma, Poole, Norouzi, Fleet, and Salimans]{ho2022imagen}
Jonathan Ho, William Chan, Chitwan Saharia, Jay Whang, Ruiqi Gao, Alexey Gritsenko, Diederik~P. Kingma, Ben Poole, Mohammad Norouzi, David~J. Fleet, and Tim Salimans.
\newblock Imagen video: High definition video generation with diffusion models, 2022.

\bibitem[Hong et~al.(2022)Hong, Lee, Jang, and Kim]{Hong2022ImprovingSQ}
Susung Hong, Gyuseong Lee, Wooseok Jang, and Seung~Wook Kim.
\newblock Improving sample quality of diffusion models using self-attention guidance.
\newblock \emph{2023 IEEE/CVF International Conference on Computer Vision (ICCV)}, pages 7428--7437, 2022.

\bibitem[Hoogeboom et~al.(2023)Hoogeboom, Heek, and Salimans]{Hoogeboom2023simpleDE}
Emiel Hoogeboom, Jonathan Heek, and Tim Salimans.
\newblock simple diffusion: End-to-end diffusion for high resolution images.
\newblock In \emph{International Conference on Machine Learning}, 2023.

\bibitem[Hu et~al.(2023)Hu, Chen, Caron, Asano, Snoek, and Ommer]{Hu2023GuidedDF}
Vincent~Tao Hu, Yunlu Chen, Mathilde Caron, Yuki~M. Asano, Cees G.~M. Snoek, and Bjorn Ommer.
\newblock Guided diffusion from self-supervised diffusion features.
\newblock \emph{ArXiv}, abs/2312.08825, 2023.

\bibitem[Hyv{{\"a}}rinen(2005)]{JMLR:v6:hyvarinen05a}
Aapo Hyv{{\"a}}rinen.
\newblock Estimation of non-normalized statistical models by score matching.
\newblock \emph{Journal of Machine Learning Research}, 6\penalty0 (24):\penalty0 695--709, 2005.

\bibitem[Karras et~al.(2022)Karras, Aittala, Aila, and Laine]{karras2022elucidating}
Tero Karras, Miika Aittala, Timo Aila, and Samuli Laine.
\newblock Elucidating the design space of diffusion-based generative models.
\newblock In \emph{Advances in Neural Information Processing Systems}, 2022.

\bibitem[Karras et~al.(2023)Karras, Aittala, Lehtinen, Hellsten, Aila, and Laine]{karras2023analyzing}
Tero Karras, Miika Aittala, Jaakko Lehtinen, Janne Hellsten, Timo Aila, and Samuli Laine.
\newblock Analyzing and improving the training dynamics of diffusion models, 2023.

\bibitem[Karras et~al.(2024)Karras, Aittala, Kynkäänniemi, Lehtinen, Aila, and Laine]{karras2024guidingdiffusionmodelbad}
Tero Karras, Miika Aittala, Tuomas Kynkäänniemi, Jaakko Lehtinen, Timo Aila, and Samuli Laine.
\newblock Guiding a diffusion model with a bad version of itself, 2024.

\bibitem[Kim et~al.(2022)Kim, Kim, Kang, and Moon]{Kim2022RefiningGP}
Dongjun Kim, Yeongmin Kim, Wanmo Kang, and Il-Chul Moon.
\newblock Refining generative process with discriminator guidance in score-based diffusion models.
\newblock In \emph{International Conference on Machine Learning}, 2022.

\bibitem[Lin et~al.(2024)Lin, Liu, Li, and Yang]{lin2023common}
Shanchuan Lin, Bingchen Liu, Jiashi Li, and Xiao Yang.
\newblock Common diffusion noise schedules and sample steps are flawed.
\newblock In \emph{Proceedings of the IEEE/CVF Winter Conference on Applications of Computer Vision (WACV)}, pages 5404--5411, 2024.

\bibitem[Lin et~al.(2014)Lin, Maire, Belongie, Hays, Perona, Ramanan, Doll{\'a}r, and Zitnick]{lin2015microsoft}
Tsung-Yi Lin, Michael Maire, Serge~J. Belongie, James Hays, Pietro Perona, Deva Ramanan, Piotr Doll{\'a}r, and C.~Lawrence Zitnick.
\newblock Microsoft coco: Common objects in context.
\newblock In \emph{European Conference on Computer Vision}, 2014.

\bibitem[Lipman et~al.(2023)Lipman, Chen, Ben-Hamu, Nickel, and Le]{lipman2023flow}
Yaron Lipman, Ricky T.~Q. Chen, Heli Ben-Hamu, Maximilian Nickel, and Matthew Le.
\newblock Flow matching for generative modeling.
\newblock In \emph{The Eleventh International Conference on Learning Representations}, 2023.

\bibitem[Liu et~al.(2024)Liu, Akhgari, Visheratin, Kamko, Xu, Shrirao, Lambert, Souza, Doshi, and Li]{liu2024playgroundv3improvingtexttoimage}
Bingchen Liu, Ehsan Akhgari, Alexander Visheratin, Aleks Kamko, Linmiao Xu, Shivam Shrirao, Chase Lambert, Joao Souza, Suhail Doshi, and Daiqing Li.
\newblock Playground v3: Improving text-to-image alignment with deep-fusion large language models, 2024.

\bibitem[Liu et~al.(2022)Liu, Gong, and Liu]{liu2022flow}
Xingchao Liu, Chengyue Gong, and Qiang Liu.
\newblock Flow straight and fast: Learning to generate and transfer data with rectified flow, 2022.

\bibitem[Lu et~al.(2022)Lu, Zhou, Bao, Chen, Li, and Zhu]{lu2022dpmsolver}
Cheng Lu, Yuhao Zhou, Fan Bao, Jianfei Chen, Chongxuan Li, and Jun Zhu.
\newblock {DPM}-solver: A fast {ODE} solver for diffusion probabilistic model sampling in around 10 steps.
\newblock In \emph{Advances in Neural Information Processing Systems}, 2022.

\bibitem[Lu et~al.(2023)Lu, Zhou, Bao, Chen, Li, and Zhu]{lu2023dpmsolver}
Cheng Lu, Yuhao Zhou, Fan Bao, Jianfei Chen, Chongxuan Li, and Jun Zhu.
\newblock Dpm-solver++: Fast solver for guided sampling of diffusion probabilistic models, 2023.

\bibitem[Odena et~al.(2016)Odena, Dumoulin, and Olah]{odena2016deconvolution}
Augustus Odena, Vincent Dumoulin, and Chris Olah.
\newblock Deconvolution and checkerboard artifacts.
\newblock \emph{Distill}, 2016.

\bibitem[Parmar et~al.(2022)Parmar, Zhang, and Zhu]{parmar2022aliased}
Gaurav Parmar, Richard Zhang, and Jun-Yan Zhu.
\newblock On aliased resizing and surprising subtleties in gan evaluation.
\newblock \emph{2022 IEEE/CVF Conference on Computer Vision and Pattern Recognition (CVPR)}, pages 11400--11410, 2022.

\bibitem[Peebles and Xie(2022)]{peebles2023scalable}
William~S. Peebles and Saining Xie.
\newblock Scalable diffusion models with transformers.
\newblock \emph{2023 IEEE/CVF International Conference on Computer Vision (ICCV)}, pages 4172--4182, 2022.

\bibitem[Pernias et~al.(2024)Pernias, Rampas, Richter, Pal, and Aubreville]{pernias2023wuerstchen}
Pablo Pernias, Dominic Rampas, Mats~Leon Richter, Christopher Pal, and Marc Aubreville.
\newblock W\"urstchen: An efficient architecture for large-scale text-to-image diffusion models.
\newblock In \emph{The Twelfth International Conference on Learning Representations}, 2024.

\bibitem[Podell et~al.(2024)Podell, English, Lacey, Blattmann, Dockhorn, M{\"u}ller, Penna, and Rombach]{podell2023sdxl}
Dustin Podell, Zion English, Kyle Lacey, Andreas Blattmann, Tim Dockhorn, Jonas M{\"u}ller, Joe Penna, and Robin Rombach.
\newblock {SDXL}: Improving latent diffusion models for high-resolution image synthesis.
\newblock In \emph{The Twelfth International Conference on Learning Representations}, 2024.

\bibitem[Poole et~al.(2023)Poole, Jain, Barron, and Mildenhall]{poole2022dreamfusion}
Ben Poole, Ajay Jain, Jonathan~T. Barron, and Ben Mildenhall.
\newblock Dreamfusion: Text-to-3d using 2d diffusion.
\newblock In \emph{The Eleventh International Conference on Learning Representations}, 2023.

\bibitem[Ramesh et~al.(2022)Ramesh, Dhariwal, Nichol, Chu, and Chen]{ramesh2022hierarchical}
Aditya Ramesh, Prafulla Dhariwal, Alex Nichol, Casey Chu, and Mark Chen.
\newblock Hierarchical text-conditional image generation with clip latents, 2022.

\bibitem[Rombach et~al.(2021)Rombach, Blattmann, Lorenz, Esser, and Ommer]{rombach2022highresolution}
Robin Rombach, A. Blattmann, Dominik Lorenz, Patrick Esser, and Bj{\"o}rn Ommer.
\newblock High-resolution image synthesis with latent diffusion models.
\newblock \emph{2022 IEEE/CVF Conference on Computer Vision and Pattern Recognition (CVPR)}, pages 10674--10685, 2021.

\bibitem[Saharia et~al.(2021)Saharia, Chan, Chang, Lee, Ho, Salimans, Fleet, and Norouzi]{Saharia2021PaletteID}
Chitwan Saharia, William Chan, Huiwen Chang, Chris~A. Lee, Jonathan Ho, Tim Salimans, David~J. Fleet, and Mohammad Norouzi.
\newblock Palette: Image-to-image diffusion models.
\newblock \emph{ACM SIGGRAPH 2022 Conference Proceedings}, 2021.

\bibitem[Saharia et~al.(2022)Saharia, Chan, Saxena, Li, Whang, Denton, Ghasemipour, Gontijo-Lopes, Ayan, Salimans, Ho, Fleet, and Norouzi]{saharia2022photorealistic}
Chitwan Saharia, William Chan, Saurabh Saxena, Lala Li, Jay Whang, Emily Denton, Seyed Kamyar~Seyed Ghasemipour, Raphael Gontijo-Lopes, Burcu~Karagol Ayan, Tim Salimans, Jonathan Ho, David~J. Fleet, and Mohammad Norouzi.
\newblock Photorealistic text-to-image diffusion models with deep language understanding.
\newblock In \emph{Advances in Neural Information Processing Systems}, 2022.

\bibitem[Salimans and Ho(2022)]{salimans2022progressive}
Tim Salimans and Jonathan Ho.
\newblock Progressive distillation for fast sampling of diffusion models.
\newblock In \emph{International Conference on Learning Representations}, 2022.

\bibitem[Salimans et~al.(2016)Salimans, Goodfellow, Zaremba, Cheung, Radford, and Chen]{salimans2016improved}
Tim Salimans, Ian~J. Goodfellow, Wojciech Zaremba, Vicki Cheung, Alec Radford, and Xi Chen.
\newblock Improved techniques for training gans.
\newblock In \emph{Advances in Neural Information Processing Systems 29: Annual Conference on Neural Information Processing Systems 2016, December 5-10, 2016, Barcelona, Spain}, pages 2226--2234, 2016.

\bibitem[Schuhmann et~al.(2022)Schuhmann, Beaumont, Vencu, Gordon, Wightman, Cherti, Coombes, Katta, Mullis, Wortsman, Schramowski, Kundurthy, Crowson, Schmidt, Kaczmarczyk, and Jitsev]{schuhmann2022laion5b}
Christoph Schuhmann, Romain Beaumont, Richard Vencu, Cade~W Gordon, Ross Wightman, Mehdi Cherti, Theo Coombes, Aarush Katta, Clayton Mullis, Mitchell Wortsman, Patrick Schramowski, Srivatsa~R Kundurthy, Katherine Crowson, Ludwig Schmidt, Robert Kaczmarczyk, and Jenia Jitsev.
\newblock {LAION}-5b: An open large-scale dataset for training next generation image-text models.
\newblock In \emph{Thirty-sixth Conference on Neural Information Processing Systems Datasets and Benchmarks Track}, 2022.

\bibitem[Shi et~al.(2024)Shi, Wang, Ye, Mai, Li, and Yang]{shi2023mvdream}
Yichun Shi, Peng Wang, Jianglong Ye, Long Mai, Kejie Li, and Xiao Yang.
\newblock {MVD}ream: Multi-view diffusion for 3d generation.
\newblock In \emph{The Twelfth International Conference on Learning Representations}, 2024.

\bibitem[Singer et~al.(2023)Singer, Polyak, Hayes, Yin, An, Zhang, Hu, Yang, Ashual, Gafni, Parikh, Gupta, and Taigman]{singer2022makeavideo}
Uriel Singer, Adam Polyak, Thomas Hayes, Xi Yin, Jie An, Songyang Zhang, Qiyuan Hu, Harry Yang, Oron Ashual, Oran Gafni, Devi Parikh, Sonal Gupta, and Yaniv Taigman.
\newblock Make-a-video: Text-to-video generation without text-video data.
\newblock In \emph{The Eleventh International Conference on Learning Representations}, 2023.

\bibitem[Sohl{-}Dickstein et~al.(2015)Sohl{-}Dickstein, Weiss, Maheswaranathan, and Ganguli]{sohldickstein2015deep}
Jascha Sohl{-}Dickstein, Eric~A. Weiss, Niru Maheswaranathan, and Surya Ganguli.
\newblock Deep unsupervised learning using nonequilibrium thermodynamics.
\newblock In \emph{Proceedings of the 32nd International Conference on Machine Learning, {ICML} 2015, Lille, France, 6-11 July 2015}, pages 2256--2265. JMLR.org, 2015.

\bibitem[Song et~al.(2021{\natexlab{a}})Song, Meng, and Ermon]{song2022denoising}
Jiaming Song, Chenlin Meng, and Stefano Ermon.
\newblock Denoising diffusion implicit models.
\newblock In \emph{International Conference on Learning Representations}, 2021{\natexlab{a}}.

\bibitem[Song et~al.(2021{\natexlab{b}})Song, Durkan, Murray, and Ermon]{Song2021MaximumLT}
Yang Song, Conor Durkan, Iain Murray, and Stefano Ermon.
\newblock Maximum likelihood training of score-based diffusion models.
\newblock In \emph{Neural Information Processing Systems}, 2021{\natexlab{b}}.

\bibitem[Song et~al.(2021{\natexlab{c}})Song, Sohl-Dickstein, Kingma, Kumar, Ermon, and Poole]{song2021scorebased}
Yang Song, Jascha Sohl-Dickstein, Diederik~P Kingma, Abhishek Kumar, Stefano Ermon, and Ben Poole.
\newblock Score-based generative modeling through stochastic differential equations.
\newblock In \emph{International Conference on Learning Representations}, 2021{\natexlab{c}}.

\bibitem[Song et~al.(2023)Song, Dhariwal, Chen, and Sutskever]{song2023consistencymodels}
Yang Song, Prafulla Dhariwal, Mark Chen, and Ilya Sutskever.
\newblock Consistency models, 2023.

\bibitem[Wang et~al.(2023)Wang, Lu, Wang, Bao, Li, Su, and Zhu]{wang2023prolificdreamer}
Zhengyi Wang, Cheng Lu, Yikai Wang, Fan Bao, Chongxuan Li, Hang Su, and Jun Zhu.
\newblock Prolificdreamer: High-fidelity and diverse text-to-3d generation with variational score distillation.
\newblock In \emph{Thirty-seventh Conference on Neural Information Processing Systems}, 2023.

\bibitem[Yan et~al.(2023)Yan, Liew, Mai, Lin, and Feng]{yan2023magicprop}
Hanshu Yan, Jun~Hao Liew, Long Mai, Shanchuan Lin, and Jiashi Feng.
\newblock Magicprop: Diffusion-based video editing via motion-aware appearance propagation, 2023.

\bibitem[Zhang et~al.(2018)Zhang, Isola, Efros, Shechtman, and Wang]{zhang2018unreasonable}
Richard Zhang, Phillip Isola, Alexei~A. Efros, Eli Shechtman, and Oliver Wang.
\newblock The unreasonable effectiveness of deep features as a perceptual metric.
\newblock In \emph{2018 {IEEE} Conference on Computer Vision and Pattern Recognition, {CVPR} 2018, Salt Lake City, UT, USA, June 18-22, 2018}, pages 586--595. {IEEE} Computer Society, 2018.

\bibitem[Zheng et~al.(2022)Zheng, Vuong, Cai, and Phung]{zheng2022movq}
Chuanxia Zheng, Long~Tung Vuong, Jianfei Cai, and Dinh Phung.
\newblock Mo{VQ}: Modulating quantized vectors for high-fidelity image generation.
\newblock In \emph{Advances in Neural Information Processing Systems}, 2022.

\bibitem[Zhou et~al.(2023)Zhou, Wang, Yan, Lv, Zhu, and Feng]{zhou2023magicvideo}
Daquan Zhou, Weimin Wang, Hanshu Yan, Weiwei Lv, Yizhe Zhu, and Jiashi Feng.
\newblock Magicvideo: Efficient video generation with latent diffusion models, 2023.

\end{thebibliography}
}

\clearpage
% \setcounter{page}{1}
% \maketitlesupplementary

\definecolor{codegreen}{rgb}{0,0.6,0}
\definecolor{codegray}{rgb}{0.5,0.5,0.5}

\lstdefinestyle{mystyle}{
    commentstyle=\color{codegreen},
    keywordstyle=\color{black},
    numberstyle=\tiny\color{codegray},
    basicstyle=\ttfamily\small,
    breakatwhitespace=false,         
    breaklines=true,                 
    captionpos=b,                    
    keepspaces=true,                 
    numbers=left,                    
    numbersep=5pt,                  
    showspaces=false,                
    showstringspaces=false,
    showtabs=false,                  
    tabsize=2,
}

\lstset{style=mystyle}

\begin{algorithm*}[b]
\begin{lstlisting}[language=Python]
# Create dataloader
dataloader = create_dataloader()

# Create model by loading from mse pretrained weights.
model = create_model(mse_pretrained=True)
optimizer = Adam(model.parameters(), lr=3e-5)

# Create perceptual model and freeze it.
perceptual_model = deepcopy(model)
perceptual_model.requires_grad_(False)
perceptual_model.eval()

# Dataloader yields image (latent) x_0, and conditional prompt c.
for x_0, c in dataloader:

    # Sample timesteps and epsilon noises.
    # Then perform forward diffusion.
    t = randint(0, 1000, size=[batch_size])
    eps = randn_like(x_0)
    x_t = forward(x_0, eps, t) # equation 1.

    # Pass through model to get v prediction.
    # Then convert v_pred to x_0_pred and eps_pred.
    v_pred = model(x_t, t, c)
    x_0_pred = to_x_0(v_pred, x_t, t) # equation 7.
    eps_pred = to_eps(v_pred, x_t, t) # equation 8.

    # Sample new timesteps.
    # Then perform forward diffusion twice.
    # One uses ground truth x_0 and eps.
    # Another uses predicted x_0_pred and eps_pred.
    tt = randint(0, 1000, size=[batch_size])
    x_tt = forward(x_0, eps, tt)
    x_tt_pred = forward(x_0_pred, eps_pred, tt)
    
    # Pass through perceptual model.
    # Get hidden feature from midblock.
    feature_real = perceptual_model(x_tt, tt, c, return_feature="midblock")
    feature_pred = perceptual_model(x_tt_pred, tt, c, return_feature="midblock")

    # Compute loss on hidden features.
    loss = mse_loss(feature_pred, feature_real)
    loss.backward()
    optimizer.step()
    optimizer.zero_grad()
\end{lstlisting}
\caption{PyTorch code snippet for self-perceptual training.}
\label{algo:self-perceptual}
\end{algorithm*}

\end{document}